%% file: main.tex
\documentclass[10pt,twocolumn,letterpaper]{article}
\usepackage[pagenumbers]{cvpr}
\input{preamble}

\definecolor{cvprblue}{rgb}{0.21,0.49,0.74}
\usepackage[pagebackref,breaklinks,colorlinks,
    citecolor=cvprblue,
]{hyperref}

\title{
Canvas-to-Image: Compositional Image Generation with Multimodal Controls
}

\usepackage{xspace}
\usepackage{multirow}

\newcommand{\methodname}{\text{Canvas-to-Image}\xspace}
\newcommand{\webpage}{\url{https://snap-research.github.io/canvas-to-image/}~}

\author{
Yusuf Dalva$^{*1,3}$\quad
Guocheng Gordon Qian$^{*\dagger1}$\quad
Maya Goldenberg$^{1}$\quad
Tsai-Shien Chen$^{1,2}$\\
Kfir Aberman$^{1}$\quad
Sergey Tulyakov$^{1}$\quad
Pinar Yanardag$^{3}$\quad
Kuan-Chieh Jackson Wang$^{1}$\\
\\
$^1$Snap Inc. \quad
$^2$UC Merced \quad
$^3$Virginia Tech\\
{\tt\small \webpage}
}
\begin{document}
\input{fig/teaser}

\maketitle

\renewcommand{\thefootnote}{\fnsymbol{footnote}}
\footnotetext[1]{Equal Contributions. $^{\dagger}$Corresponding author.}
\renewcommand{\thefootnote}{\arabic{footnote}}

\input{sec/0_abstract}    
\input{sec/1_intro}
\input{sec/2_related_work}

\input{sec/3_method}

\input{sec/4_experiments}
\input{sec/ablation}
\input{sec/5_conclusion}

{
    \small
    \bibliographystyle{ieeenat_fullname}
    \bibliography{main}
} 
\input{sec/X_suppl}

\end{document}

%% file: preamble.tex
\usepackage{bm}

\newcommand{\figlabel}{Fig.\xspace}

\newcommand{\inlinesection}[1]{\noindent\textbf{#1.}}

\newcommand{\supp}{\textit{Appendix}\xspace}

\definecolor{Highlight}{HTML}{39b54a}  %

\usepackage{amssymb}
\usepackage{pifont}

\usepackage{tcolorbox}
\tcbuselibrary{skins, breakable}

\usepackage{algorithm}
\usepackage{listings}
\usepackage{etoolbox}
\makeatletter
\AfterEndEnvironment{algorithm}{\let\@algcomment\relax}
\AtEndEnvironment{algorithm}{\kern2pt\hrule\relax\vskip3pt\@algcomment}
\let\@algcomment\relax
\newcommand\algcomment[1]{\def\@algcomment{\footnotesize#1}}
\renewcommand\fs@ruled{\def\@fs@cfont{\bfseries}\let\@fs@capt\floatc@ruled
  \def\@fs@pre{\hrule height.8pt depth0pt \kern2pt}%
  \def\@fs@post{}%
  \def\@fs@mid{\kern2pt\hrule\kern2pt}%
  \let\@fs@iftopcapt\iftrue}
\makeatother
\newcommand{\cmmnt}[1]{}

\usepackage[normalem]{ulem}

%% file: fig/teaser.tex
\twocolumn[{
\maketitle
\begin{center}
    \captionsetup{type=figure}
    \vspace{-1em}
\newcommand{\imwidth}{1\textwidth}
\begin{tabular}{@{}c@{}}
\parbox{\imwidth}{ \centering \includegraphics[width=\imwidth, width=\linewidth]{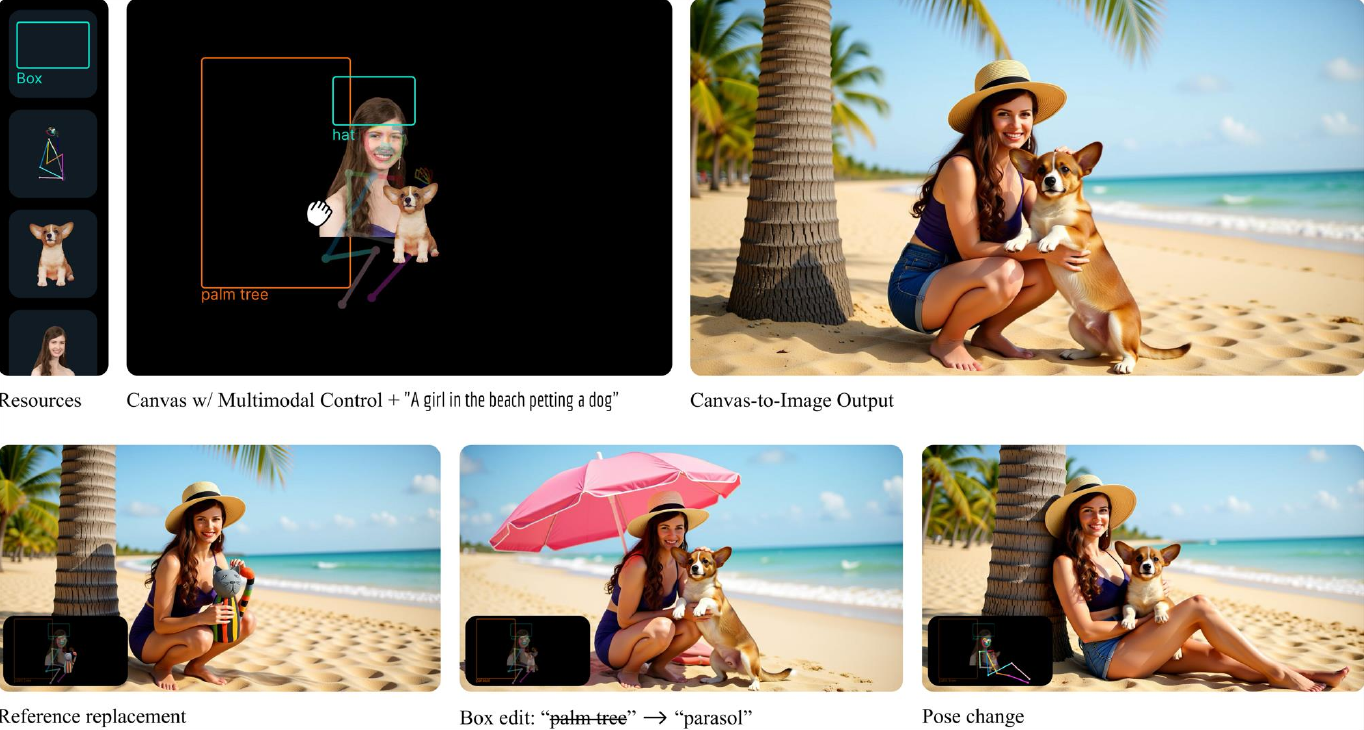}}

\\

\vspace{1em}
\end{tabular}
    \vspace{-2.5em}
    \captionof{figure}{\textbf{\methodname{}} enables compositional control for text-to-image generation through a unified Multi-Task Canvas framework. The canvas serves as a flexible visual interface that guides image synthesis by supporting diverse guiding signals, including spatially positioned subjects, pose signals, bounding boxes, and text annotations. 
    } 
    \label{fig:teaser}
\end{center}
}]

%% file: sec/0_abstract.tex
\begin{abstract}
While modern diffusion models excel at generating high-quality and diverse images, they still struggle with high-fidelity compositional and multimodal control, particularly when users simultaneously specify text prompts, subject references, spatial arrangements, pose constraints, and layout annotations. We introduce Canvas-to-Image, a unified framework that consolidates these heterogeneous controls into a single canvas interface, enabling users to generate images that faithfully reflect their intent.
Our key idea is to encode diverse control signals into a single composite canvas image that the model can directly interpret for integrated visual-spatial reasoning. We further curate a suite of multi-task datasets and propose a Multi-Task Canvas Training strategy that optimizes the diffusion model to jointly understand and integrate heterogeneous controls into text-to-image generation within a unified learning paradigm. This joint training enables Canvas-to-Image to reason across multiple control modalities rather than relying on task-specific heuristics, and it generalizes well to multi-control scenarios during inference.
Extensive experiments show that Canvas-to-Image significantly outperforms state-of-the-art methods in identity preservation and control adherence across challenging benchmarks, including multi-person composition, pose-controlled composition, layout-constrained generation, and multi-control generation.
\end{abstract}

%% file: sec/1_intro.tex
\section{Introduction}
\label{sec:intro}

Recent advances in large-scale diffusion models~\citep{rombach2022high,podell2023sdxl,esser2024scaling} have substantially improved the realism and diversity of synthesized imagery. However, these models remain inherently stochastic and provide limited flexibility when users wish to control multiple aspects of image generation simultaneously. This limitation is particularly consequential in creative and design-oriented applications, such as digital art and content creation, where users often need to coordinate several types of control signals, such as spatial layouts, subject references, pose constraints, etc.

We introduce \textbf{\methodname{}}, a framework that enables \textit{heterogeneous compositional control} over diffusion-based image generation through a unified canvas representation. As illustrated in \figlabel\ref{fig:teaser}, our approach allows users to combine diverse forms of input within a single interface: subjects and objects can be positioned, resized, rotated, and posed; bounding boxes with descriptive tags can define subjects with spatial constraints; and pose overlays~\citep{cao2019openpose} can specify body configurations. This flexible, multi-modal interaction design enables users to guide the generation process using complementary controls that collaboratively define both semantics and composition.

Achieving unified multi-control generation remains highly challenging, and no existing model can handle all the aforementioned controls simultaneously. Existing control mechanisms~\citep{mou2023t2i,li2023gligen,zhang2023adding} typically address \textit{isolated} aspects of compositional image synthesis, e.g. spatial layouts or pose constraints, but \textit{fail to handle multiple controls within a single input}. The core difficulty lies in reconciling heterogeneous inputs that differ in both structure and semantics, including subject references, bounding boxes, and textual tags, while training a model capable of jointly interpreting and balancing these signals.

Consequently, prior works supporting subject injection~\citep{ye2023ip-adapter,qian2025composeme,han2024emma} usually lack spatial control, whereas layout-guided methods~\citep{li2023gligen,zhang2025creatidesign} cannot incorporate specific poses or subjects. Recent methods such as StoryMaker~\citep{StoryMaker} and ID-Patch~\citep{ID-Patch} demonstrate both subject insertion and spatial control, but rely on complex module combinations, such as ControlNet~\citep{flux_controlnet} and IP-Adapter~\citep{ye2023ip-adapter}, which introduce additional complexity, are limited to face injection, lack bounding-box support, and generalize poorly.

To address these challenges, we propose three key innovations. \textbf{First}, we introduce the \textbf{Multi-Task Canvas}, a unified input representation that consolidates diverse control modalities, including background composition, subject insertion, bounding-box layouts, and pose guidance, into a single composite RGB image. This \textit{canvas} serves as a generalized visual interface where all control elements are expressed in a common pixel space, allowing the model to interpret multimodal guidance without extra modules or architectural changes. \textbf{Second}, we curate a comprehensive multi-task dataset that aligns these heterogeneous controls with corresponding target images, supporting consistent joint supervision across tasks. \textbf{Third}, we design a \textbf{Multi-Task Canvas Training} framework that fine-tunes the diffusion model to reason across tasks collectively, learning shared semantics and spatial dependencies among different control types. Importantly, we observe that once trained on this multi-task canvas, the model generalizes naturally to \textit{multi-control} scenarios at inference time, even when combinations of controls were not seen together during training.
We summarize our \textbf{contributions} as follows:
\begin{itemize}[leftmargin=*,noitemsep,topsep=0pt,parsep=0pt,partopsep=0pt]
    \item \textbf{Unified canvas framework:} A generalized \textit{Multi-Task Canvas} representation that consolidates heterogeneous controls into a single canvas-to-image formulation (\figlabel\ref{fig:pipeline}), enabling coherent reasoning across modalities.

    \item \textbf{Multi-task datasets and training:} We curate comprehensive multi-task datasets covering diverse control modalities and propose a unified \textit{Multi-Task Canvas Training} framework that fine-tunes the diffusion model jointly across these tasks. Experiments reveal joint training enables mixed controls in inference time.

    \item \textbf{Comprehensive evaluation:} Extensive experiments on challenging benchmarks demonstrate clear improvements in identity preservation and control adherence compared to existing methods. Ablations confirm that our unified multi-task design is key to achieving flexible and coherent heterogeneous control.
\end{itemize}

%% file: sec/2_related_work.tex
\section{Related Work}
\label{sec:related_work}

\input{fig/pipeline}

\inlinesection{Diffusion Models for Image Synthesis}
Diffusion models~\citep{ho2020denoising,song2020denoising} are the dominant paradigm for high-fidelity image synthesis. Text-to-image models~\citep{ramesh2022hierarchical,saharia2022photorealistic,rombach2022high} use large-scale text-image pairs for open-vocabulary generation. Diffusion transformers~\citep{peebles2023scalable,esser2024scaling,fluxkontext} have further improved quality and scalability. Emerging multimodal models~\citep{deng2025emerging, wu2025qwen} integrate MLLMs with diffusion models for higher quality and better prompt following. While impressive, these models still struggle with fine-grained, multi-constraint compositional control. Our work builds on a pretrained diffusion model, introducing a unified canvas interface and multi-task training strategy to enable comprehensive compositional control.

\inlinesection{Personalization in Image Generation}
Personalization methods generate specific subjects or identities in novel contexts. Early approaches~\citep{gal2022image,ruiz2023dreambooth,custom-diffusion} require per-concept fine-tuning. Adapter-based solutions~\citep{ye2023ip-adapter,wang2024instantid,PuLID,Omni-ID,patashnik2025nested,goyal2025preventing} improve efficiency by keeping the base model frozen and injecting subject-specific representations. Multi-concept personalization remains challenging: optimization-based methods~\citep{avrahami2023break,tokenverse,po2023orthogonal,kong2024omg,dalva2025lorashop} demand explicit concept disentanglement, while optimization-free approaches~\citep{xiao2024fastcomposer,wang2024moa,han2024emma,qian2025composeme,chen2025multi} concatenate embeddings at the cost of linear complexity growth. Beyond these scalability and flexibility limitations, most personalization methods focus solely on reference injection~\cite{qian2025layercomposer,he2024uniportrait}. A true creative design process requires handling multiple controls simultaneously. \methodname{} addresses the scalability and flexible control challenges with a unified single canvas that maintains a constant computation cost, providing a foundation for such a multi-control, multi-subject personalization framework.

\inlinesection{Compositional Control in Generation}
Providing fine-grained compositional control remains a challenge, as existing mechanisms typically address isolated tasks. For instance, models like ControlNet~\citep{zhang2023adding} and T2I-Adapter~\citep{mou2023t2i} use structural cues like pose skeletons or depth maps to specify body configurations. Another line of work targets spatial layout control. Methods such as GLIGEN~\citep{li2023gligen}, LayoutDiffusion~\citep{zheng2023layoutdiffusion}, and CreatiDesign~\citep{zhang2025creatidesign} finetune the generator to interpret bounding boxes or segmentation masks. Unifying these heterogeneous controls is highly challenging, particularly with identity constraints for personalization. Methods supporting subject injection often lack fine-grained spatial control, while layout-guided methods cannot incorporate specific poses or subject identities. Recent attempts at unification, such as StoryMaker~\citep{StoryMaker} and ID-Patch~\citep{ID-Patch}, rely on complex combinations of separate modules (e.g., ControlNet with IP-Adapter) and are limited to single-type control. \methodname{} addresses this gap by reformulating diverse control types into a single ``visual canvas''. Instead of relying on task-specific heuristics, our unified canvas supports spatial layouts, pose guidance, and subject appearance injection within one coherent interface, enabling the model to reason across modalities collectively.

%% file: fig/pipeline.tex
\begin{figure*}[!t]
    \centering
    \includegraphics[width=\linewidth]{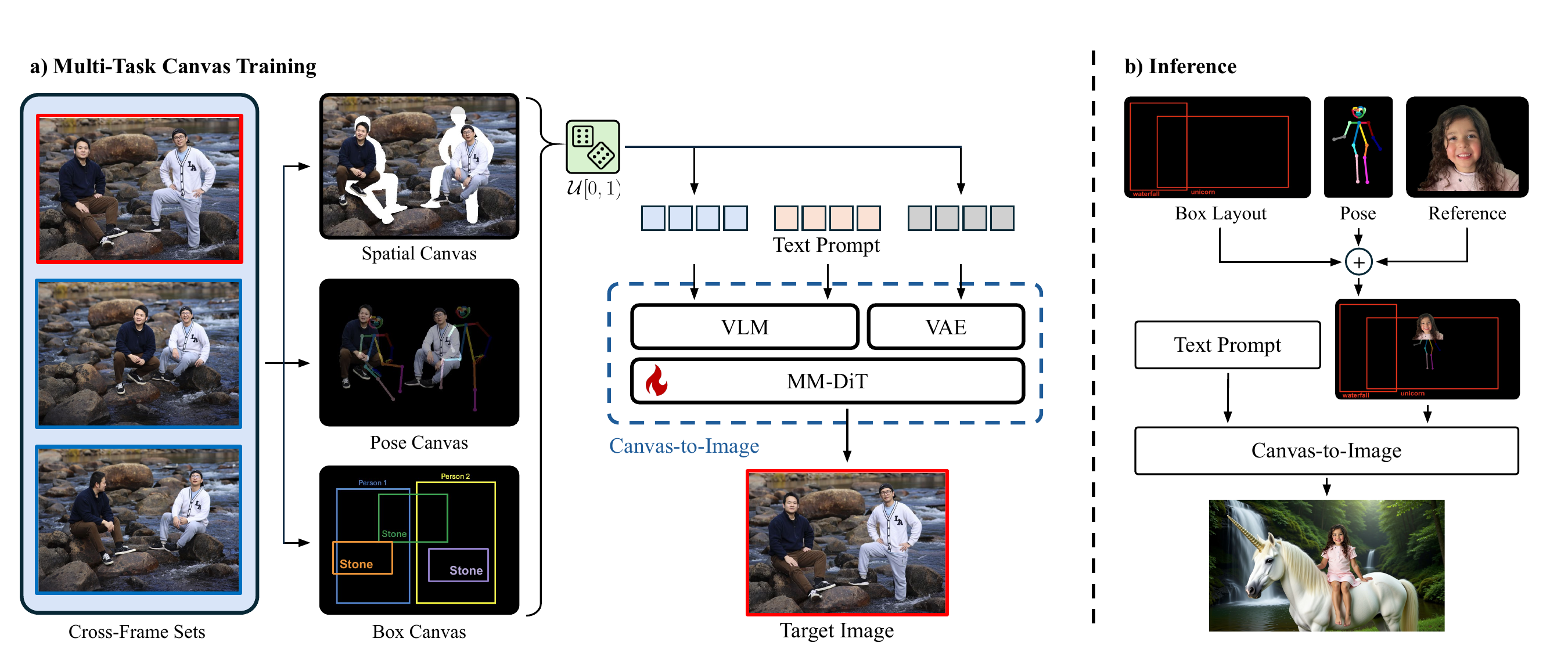}
    \caption{\textbf{Overview of \methodname{} framework.} 
 (a) \textbf{Multi-Task Canvas Training.} We reformulate heterogeneous control tasks: spatial composition, pose guidance, and layout-constrained generation into a single \textit{canvas-to-image} formulation. Each training step samples one type of canvas (Spatial, Pose, or Box), where the target frame serves as supervision. All control signals are encoded as RGB canvases interpretable by the Vision-Language Model (VLM) for unified visual–spatial reasoning. The Multi-Modal DiT (MM-DiT) receives VLM embeddings, VAE latents, and noisy latents to predict the velocity for flow matching.
(b) \textbf{Inference.} Although trained on single-control samples, the model generalizes to multi-control compositions, jointly leveraging pose, layout, and reference cues within a single generation process. This enables coherent multi-control reasoning without task-specific retraining.}
    \label{fig:pipeline}
\end{figure*}

%% file: sec/3_method.tex
\section{Methodology}
\label{sec:method}

\methodname{} is a unified framework for multi-modal, compositionally controlled image synthesis. The model takes as input a generalized Multi-Task Canvas, which is a single RGB image used to encode heterogeneous user controls. These controls include subject identities for personalization, spatial layouts, human poses, or bounding boxes. The Multi-Task Canvas formulation (Sec.~\ref{sec:multitask-canvas}) enables the diffusion model to interpret these diverse control modalities, all unified within this single image format, within a consistent training setup. Each canvas variant teaches the model a different type of compositional reasoning, from using subject references for personalization to applying fine-grained structural guidance. The underlying VLM–Diffusion architecture and multi-task training strategy (Sec.~\ref{sec:multitask-training}) jointly optimize the model across all control types. This design enables \methodname{} to generalize to multi-control scenarios at inference, combining conditions not seen together during training while maintaining precise, controllable synthesis.

\subsection{Multi-Task Canvas}
\label{sec:multitask-canvas}

Our core contribution is the introduction of a Multi-Task Canvas that generalizes different complex compositional tasks into a shared input format: a single RGB image. This ``visual canvas'' serves as a flexible, multi-modal format that unifies diverse compositional inputs. We generate our canvas variants, which serve as different ways of expressing a composition, from data sources appropriate for each task. These variants are designed to be interpreted as distinct control types. For example, a Spatial Canvas provides a literal, pixel-based composition, while a Pose Canvas provides an abstract, structural one. \methodname{} is built upon three primary canvas variants:

\inlinesection{Spatial Canvas}
The first variant trains the model to render a complete scene based on an explicit composition, as depicted in \cref{fig:pipeline} as ``Spatial Canvas''. This input canvas is a composite RGB image created by visually pasting segmented cutouts of subjects (e.g., $I_{\text{subject\_1}}, I_{\text{subject\_2}}$) at their desired locations on a masked background. This canvas is constructed using Cross-Frame Sets (\cref{fig:pipeline} left), which allows for pairing subjects and backgrounds drawn in a cross-frame manner. This strategy is crucial for the methodology as it avoids the copy-pasting artifacts common in simpler composition methods. This canvas enables multi-subject personalization as a compositional control, where users can place and resize reference subjects to guide the generation.

\inlinesection{Pose Canvas}
This task enhances the Spatial Canvas by providing a strong visual constraint for articulation. We overlay a ground-truth pose skeleton (\eg, from \cite{cao2019openpose}) onto the Spatial Canvas using a specific transparency factor, as shown in \cref{fig:pipeline} as "Pose Canvas". This semi-transparent overlay is a key design choice: the pose skeleton remains clearly recognizable as a structural guide, while the visual identity from the underlying subject segments (when present) can still be recovered and interpreted by the model. In this canvas, the subject segments themselves are randomly dropped during training, i.e., there are cases with only poses in the empty canvas to guide the pose. This is designed to support pose control as an independent modality in inference, even without reference injection.

\inlinesection{Box Canvas}
This task trains the model to interpret explicit layout specifications through bounding boxes with textual annotations directly onto the canvas. Each box contains a textual identifier (e.g., "Person 1", "Person 2", "Stone" in \cref{fig:pipeline}) that specifies which subject should appear in that spatial region and their sizes. The person identifier is ordered from left to right. Such a ``Box Canvas'' supports simple spatial control with text annotations without reference images as in previous two canvas variants.

 By training the model on these distinct, single-task canvas types, the framework learns a robust and generalizable policy for each control. Interestingly, this enables the model to generalize beyond single-task learning, allowing for the simultaneous execution of these distinct control signals at inference time even in combinations not encountered during training, as shown in \cref{fig:pipeline}(b).

\subsection{Model and Multi-Task Training}
\label{sec:multitask-training}

As illustrated in Fig. \ref{fig:pipeline}, \methodname{} builds upon a VLM–Diffusion architecture. The Vision-Language Model (VLM) encodes the unified canvas into a tokenized representation. This representation is concatenated with the VAE latents of the canvas and provided to the diffusion model as conditional inputs, along with the text prompt embedding and the noisy latents. The model is optimized using a task-aware Flow-Matching Loss:

\begin{equation}
    \label{eqn:flow}
    \mathcal{L}_{\text{flow}} 
    = \mathbb{E}_{\substack{x_0, x_1, t, \\ h, c}} 
    \left[
        \left\|
            v_{\theta}\big(x_t, t, [h; c]\big) 
            - (x_0 - x_1)
        \right\|_2^2
    \right],
\end{equation}
where $x_0$ is the target latent, $x_1$ is the noise latent, and $x_t$ is the interpolated latent. $h$ represents the input condition, which is itself a concatenation of the VLM embeddings (derived from the canvas and text prompt) and the VAE latents (derived from the same canvas). $c$ represents the task indicator that specifies the current control type. The network $v_{\theta}$ predicts the target velocity $v_t = x_0 - x_1$.

\methodname{} adopts a unified \textit{Multi-Task Canvas} formulation, in which each training step samples one type of canvas as the input condition (\eg, \textit{Spatial}, \textit{Pose}, \textit{Box}). Training on this diverse, multi-task curriculum enables the model to learn decoupled, generalizable representations for each control type. Consequently, the model can execute a combination of these controls at inference time (e.g., a mixed canvas with both pose skeletons and layout boxes) despite having never seen such a combination during training. This emergent generalization from single-task learning to multi-task application is a key property of the proposed framework.
To prevent task interference, we introduce a task indicator prompt---a short textual token (e.g., ``[Spatial]'', ``[Pose]'' or ``[Box]'') prepended to the user prompt. This indicator ($c$), which is necessary because our different canvas types represent different control meanings, disambiguates the task context and prevents mode blending. Ablation studies (Sec. 4.3) demonstrate the effectiveness of our multi-task training strategy on performing these control tasks compositionally at inference time.

%% file: sec/4_experiments.tex
\input{fig/4p_bench}

\section{Experiments}
\label{sec:experiments}

\input{fig/4p-pose}

\subsection{Experiment Details}
\inlinesection{Implementation}
We build upon Qwen-Image-Edit~\citep{wu2025qwen} as our base architecture. The input canvas image and text prompt are first processed by the VLM to extract semantic embeddings, while the canvas image is also simultaneously encoded by the VAE into latents. These VLM embeddings, VAE latents, and noisy latents are concatenated and fed into the diffusion, which predicts the velocity for denoising. During training, we fine-tune the attention, image modulation, and text modulation layers in each block using LoRA~\cite{hu2022lora} with a rank of 128. Note the feed-forward layers are frozen, as we find it is important to preserve the prior image quality of the pretrained model. Optimization is performed with AdamW~\cite{AdamW}, using a learning rate of $5\times10^{-5}$ and an effective batch size of 32. The model is trained for 200K steps on 32 NVIDIA A100 GPUs.

\inlinesection{Dataset}
\label{sec:data}
Our training is constructed from two primary data sources.  The \textit{Spatial Canvas} and \textit{Pose Canvas} variants are derived from a large-scale internal, human-centric dataset containing 6M cross-frame images from 1M unique identities. This dataset enables flexible composition sampling for our \textit{Multi-Task Canvas} formulation, for example, pairing subjects and backgrounds drawn in a cross-frame manner to avoid copy-pasting artifacts. See \supp for details. 
For the \textit{Box Canvas}, we extend the internal data with bounding box annotations from the external CreatiDesign dataset~\cite{zhang2025creatidesign}, which provides a large-scale corpus of images annotated with boxes and named entities. 
During training, we sample each task type and its dataset with an uniform distribution for a balanced multi-task supervision.

\inlinesection{Benchmarks}
We benchmark our method against several baselines, including the base model Qwen-Image-Edit~\cite{wu2025qwen}, the state-of-the-art commercial editing model Gemini~2.5~Flash~Image (also known as Nano-Banana)~\cite{comanici2025gemini}, and other most recent related work such as CreatiDesign~\cite{zhang2025creatidesign} and Overlay Kontext~\cite{ilkerzgi2025overlay} in corresponding benchmarks. 
For a fair and direct comparison of unified-interface methods, our main paper evaluates baselines that also operate on a single image input. We provide an extended comparison against other methods such as ID-Patch~\cite{ID-Patch} in \supp.
Evaluations are conducted across four distinct benchmarks: \textit{(i)} 4P Composition via the \textit{Spatial Canvas}, \textit{(ii)} Pose-Overlaid 4P Composition via the \textit{Pose Canvas}, \textit{(iii)} the \textit{Layout-Guided Composition} benchmark via the \textit{Box Canvas}, and \textit{(iv)} our proposed \textit{Multi-Control Benchmark}, which is curated from the CreatiDesign benchmark \cite{zhang2025creatidesign} containing humans in prompts and augmented with our Spatial and Pose Canvas for reference subject injection and pose controlling. See \supp for more details.

\inlinesection{Metrics}
We report ArcFace ID Similarity~\cite{ArcFace} for identity preservation, HPSv3~\cite{ma2025hpsv3} for image quality, VQAScore~\cite{VQAScore} for text-image alignment. In addition, to assess the fidelity w.r.t. applied control (e.g. identity, pose, box), we introduce a Control-QA score (evaluated by an LLM). For each control, Control-QA assesses each image between a score of 1-to-5, depending on how aligned each generation to the given set of control combinations. We provide details of the Control-QA in \supp. A comprehensive user study further validating these results is also provided in \supp.

\input{fig/creatidesign}

\input{fig/multicontrol}

\input{table/quantitative}

\subsection{Qualitative and Quantitative Results}
\label{sec:results}

\input{fig/ablation}

We present qualitative comparisons across the four benchmark setups in Figures~\ref{fig:4p-bench}-\ref{fig:mixed-control}. 
In the \textit{4P Composition} benchmark (\cref{fig:4p-bench}), \methodname{} demonstrates superior identity preservation and spatial alignment when composing multiple personalized subjects, outperforming state-of-the-art baselines including Qwen-Image-Edit~\cite{wu2025qwen}, the commercial model Nano-Banana~\cite{comanici2025gemini}, and Overlay~Kontext~\cite{ilkerzgi2025overlay}, which is trained upon FLUX Kontext~\cite{fluxkontext}. 
Nano-Banana consistently produces copy-pasted human segments, an observation supported by the quantitative results in \cref{tab:main_results}.  Such artifacts may occur because closed-source models such as Nano-Banana~\cite{comanici2025gemini} are likely not trained with segment-like inputs, which are explicitly incorporated in our canvas-based training. Overlay~Kontext and Qwen-Image-Edit~\cite{wu2025qwen} also fail to preserve subject identities (\eg, 1$^{\text{st}}$ row 4$^{\text{th}}$ ID, 2$^{\text{nd}}$ row 3$^{\text{rd}}$ ID, and 3$^{\text{rd}}$ row 4$^{\text{th}}$ ID), a weakness reflected in their low ArcFace scores in \cref{tab:main_results}. 

In the benchmark with extra overlaid poses (\cref{fig:pose-bench}), \methodname{} is the only method that accurately follows the target poses (``Pose Prior'' column) while maintaining high identity fidelity and visual realism, substantially outperforming baselines~\cite{comanici2025gemini,wu2025qwen}. For the \textit{Layout-Guided Composition} benchmark (\cref{fig:creatidesign}), \methodname{} produces semantically coherent compositions that adhere to the box constraints, whereas Nano-Banana and Qwen-Image-Edit often ignore structural signals or suffer from annotation rendering artifacts. Notably, \methodname{} also surpasses the dedicated state-of-the-art model CreatiDesign~\cite{zhang2025creatidesign}, which was trained specifically for this task in the training set of CreatiDesign evaluation benchmark. 

Finally, on the \textit{Multi-Control Benchmark} (see \cref{fig:mixed-control}), where identity preservation, pose guidance, and box annotations must be satisfied jointly, our model achieves the highest compositional fidelity. It integrates reference subjects and multiple control cues seamlessly, while baselines~\cite{comanici2025gemini,wu2025qwen} often produce artifacts or fail to satisfy all input constraints. Quantitatively, \cref{tab:main_results} validates the effectiveness of our unified framework. The balanced performance across control adherence and identity preservation confirms that encoding heterogeneous signals into
a single canvas successfully enables the simultaneous execution of spatial, pose, and identity constraints. 

We highlight that all results across benchmarks are generated by the same unified \methodname{} model, demonstrating its strong generalization from single-control training samples to complex control scenarios at inference. 

%% file: fig/4p_bench.tex
\begin{figure*}[!t]
    \centering
    \includegraphics[width=\linewidth]{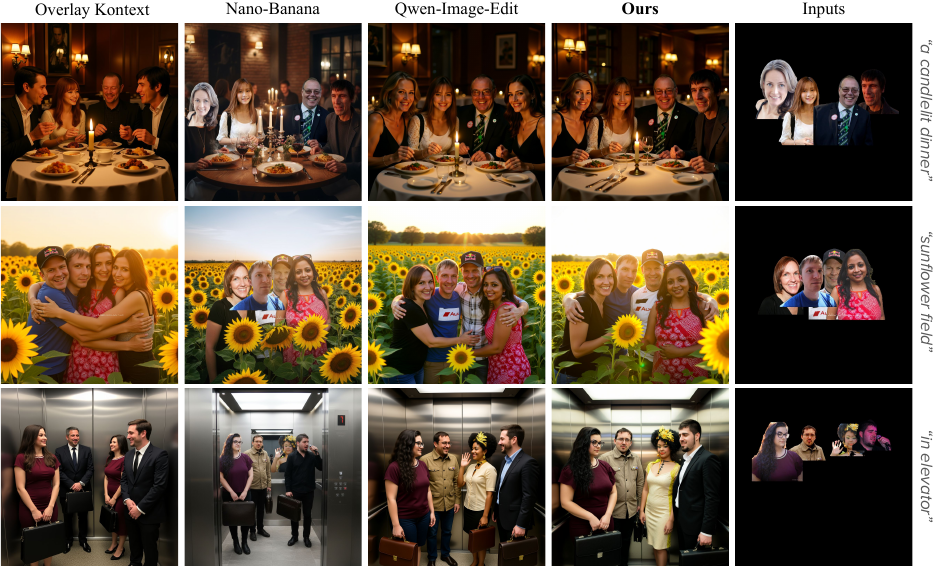}
    \caption{\textbf{Qualitative Comparisons on 4P Composition Benchmark.}
    Under the \textit{Spatial Canvas} setup, our \methodname{} achieves the highest identity preservation for multi-subject insertion while respecting the spatial placement of each subject segment. FLUX Kontext~\cite{fluxkontext}-based approach~\cite{ilkerzgi2025overlay} fails to preserve identity, whereas NanoBanana~\cite{comanici2025gemini} consistently exhibits copy-pasting artifacts. Compared to our base model, Qwen-Image-Edit~\cite{wu2025qwen}, our method maintains similar image quality but demonstrates significantly stronger identity preservation.
    }
    \label{fig:4p-bench}
\end{figure*}

%% file: fig/4p-pose.tex
\begin{figure*}[!t]
    \centering
    \includegraphics[width=\linewidth]{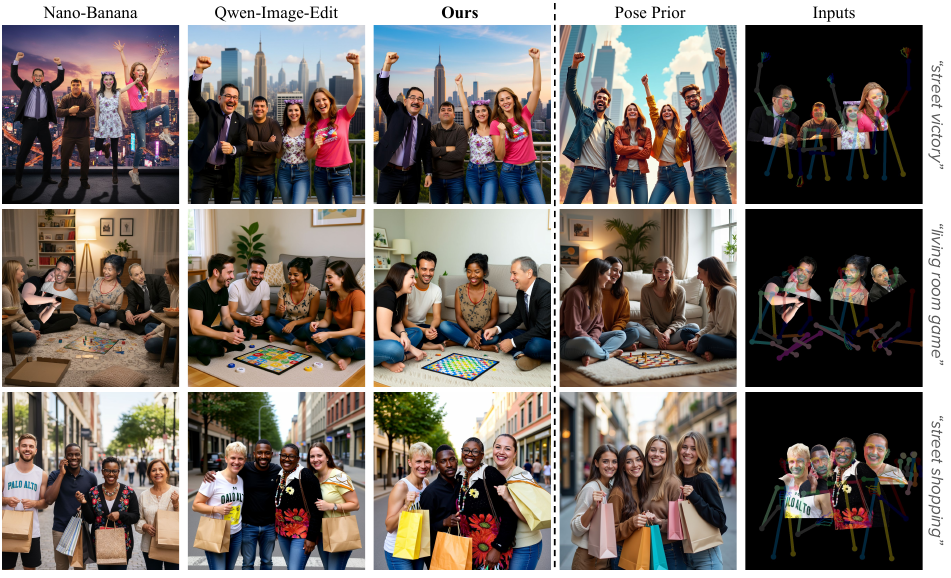}
    \caption{\textbf{Qualitative Comparisons on Pose-Overlaid 4P Composition Benchmark.}
    Our \methodname{} achieves the highest identity preservation and most accurate pose alignment. 
    Note how \methodname{} closely follows the target poses defined in the prior generated by FLUX-Dev~\cite{FLUX} (``Pose Prior'' column), while maintaining subject identities more faithfully than the baselines.}
    \label{fig:pose-bench}
\end{figure*}

%% file: fig/creatidesign.tex
\begin{figure*}[!t]
    \centering
    \includegraphics[width=\linewidth]{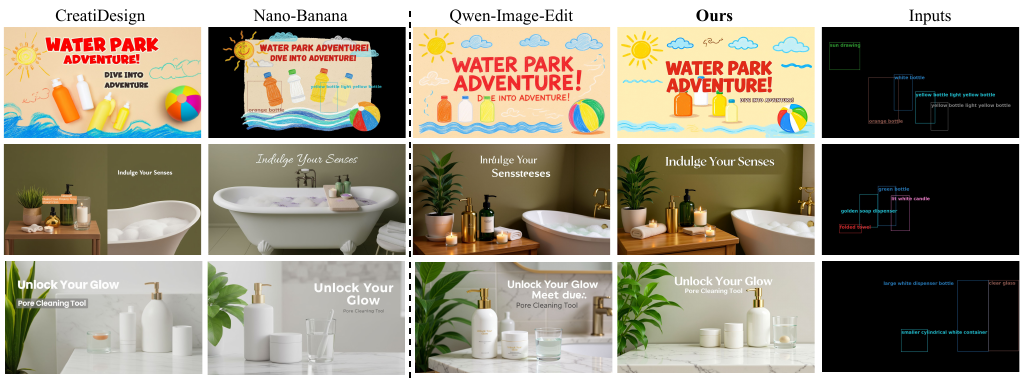}
    \caption{\textbf{Qualitative Comparisons on the Layout-Guided Composition Benchmark.} Under the \textit{Box Canvas} setup, our \methodname{} achieves the highest fidelity in spatial layout control, even compared to the state-of-the-art CreatiDesign~\cite{zhang2025creatidesign} model trained for this task. Nano Banana~\cite{comanici2025gemini}, while demonstrating good image quality, does not adhere to the bounding boxes as closely as our model. Compared to our base model Qwen-Image-Edit~\cite{wu2025qwen}, we achieve the same level of image quality but significantly stronger spatial condition alignment.
    }
    \label{fig:creatidesign}
\end{figure*}

%% file: fig/multicontrol.tex
\begin{figure*}[!t]
    \centering
    \includegraphics[width=\linewidth]{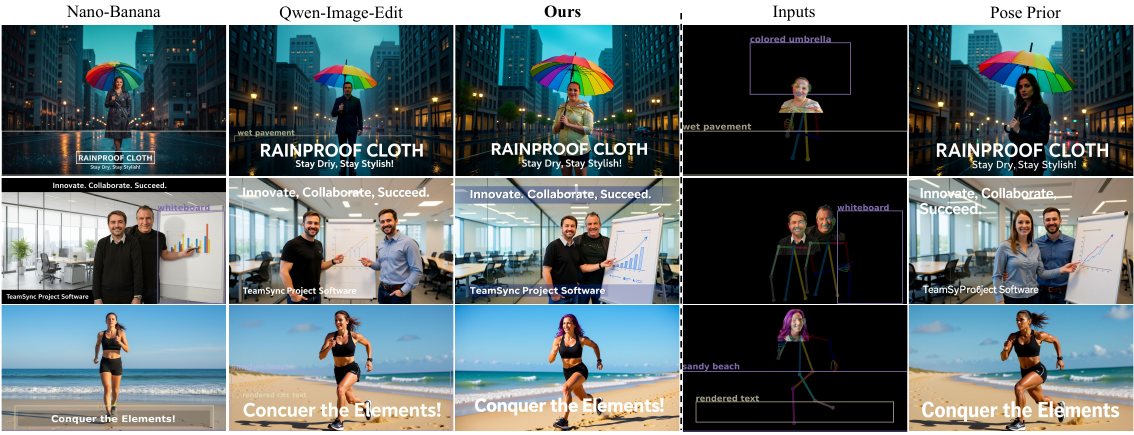}
    \caption{\textbf{Qualitative Comparisons on the Multi-Control Composition Benchmark.}
    We compare \methodname{} with state-of-the-art baselines under inputs containing multiple heterogeneous control signals. Existing methods~\cite{comanici2025gemini,wu2025qwen} fail to simultaneously satisfy all conditions, often neglecting spatial, pose, or identity constraints. In contrast, \methodname{} accurately adheres to the bounding boxes for spatial placement, respects pose and interaction cues from overlaid skeletons, and maintains strong identity fidelity of the reference identity images across multi-control inputs.}
    \label{fig:mixed-control}
\end{figure*}

%% file: table/quantitative.tex
\begin{table}[t]
\centering
\caption{\textbf{Quantitative Comparison} of our method against baselines across four different control tasks. We report ArcFace ID Similarity~\cite{ArcFace} for identity preservation, HPSv3~\cite{ma2025hpsv3} for image quality, VQAScore~\cite{VQAScore} for text--image alignment, and Control-QA for control adherence. The best results for each task are highlighted in \textbf{bold}, where the second best is highlighted as \underline{underlined}.
}

\label{tab:main_results}
\resizebox{\linewidth}{!}{%
\begin{tabular}{lcccc} 
    \toprule
    \textbf{Method} & \textbf{ArcFace}~$\uparrow$ & \textbf{HPSv3}~$\uparrow$ & \textbf{VQAScore}~$\uparrow$ & \textbf{Control-QA}~$\uparrow$ \\
    \midrule
    
    \multicolumn{5}{c}{4P Composition} \\
    Qwen-Image-Edit~\cite{wu2025qwen}& 0.258 & 13.136 & 0.890 & 3.688 \\
    Nano Banana~\cite{comanici2025gemini} & 0.434 & 10.386 & 0.826 & 3.875 \\
    Overlay Kontext~\cite{ilkerzgi2025overlay} & 0.171 & 12.693 & 0.879 & 2.000 \\
    \textbf{Ours} & \textbf{0.592} & \textbf{13.230} & \textbf{0.901} & \textbf{4.000} \\
    \midrule

    \multicolumn{5}{c}{Pose Guided 4P Composition} \\
    Qwen-Image-Edit~\cite{wu2025qwen} & 0.153 & \textbf{12.940} & 0.890 & 4.031 \\
    Nano Banana~\cite{comanici2025gemini} & 0.262 & 9.973 & 0.861 & 3.438 \\
    \textbf{Ours} & \textbf{0.300} & \underline{12.899} & \textbf{0.897} & \textbf{4.469} \\
    \midrule

    \multicolumn{5}{c}{Layout-Guided Composition} \\
    Qwen-Image-Edit~\cite{wu2025qwen}& - & 10.852 & 0.924 & 3.813 \\
    Nano Banana~\cite{comanici2025gemini} & - & 10.269 & 0.917 & 3.750 \\
    CreatiDesign~\cite{zhang2025creatidesign} & - & 9.790 & 0.923 & \textbf{4.844} \\
    \textbf{Ours} & - & \textbf{10.874} & \textbf{0.935} & \textbf{4.844} \\
    \midrule

    \multicolumn{5}{c}{Multi-Control Composition} \\
    Qwen-Image-Edit~\cite{wu2025qwen}& 0.204 & \textbf{12.251} & 0.903 & 3.575 \\
    Nano Banana~\cite{comanici2025gemini} & 0.356 & 11.370 & 0.873 & 3.625 \\
    \textbf{Ours} & \textbf{0.375} & \underline{12.044} & \textbf{0.906} & \textbf{4.281} \\
    \bottomrule
\end{tabular}
}
\vspace{-2em}
\end{table}

%% file: fig/ablation.tex
\begin{figure*}[!t]
    \centering
    \includegraphics[width=\linewidth]{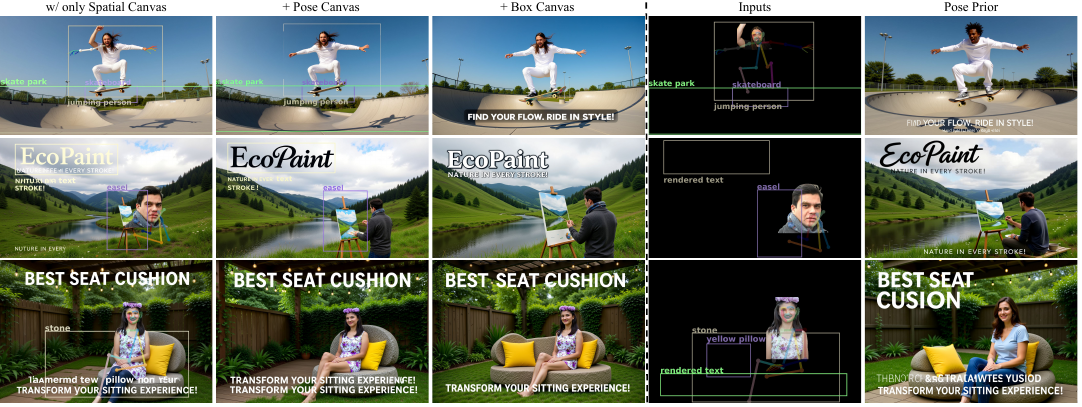}
    \caption{\textbf{Qualitative Ablation for Multi-Task Canvas Training.}
    Starting from training with only the {Spatial Canvas}, the model struggles to follow pose and bounding-box annotations. As we incrementally add the {Pose Canvas} and {Box Canvas} tasks, the model progressively learns to respect these additional controls. The final model effectively handles complex multi-control inputs. Notably, during training, each sample contains only a single control type, yet the model exhibits strong generalization to multi-control scenarios at inference. }
    \label{fig:ablation-task}
\end{figure*}

%% file: sec/ablation.tex
\input{table/ablation}
\subsection{Ablation Studies}
\label{sec:ablations}

We conduct ablation studies to evaluate the effectiveness of our \textbf{Multi-Task Canvas Training} on the \textit{Multi-Control Benchmark}. We start with a baseline model trained \textit{only} on the \textit{Spatial Canvas} and then progressively add the \textit{Pose Canvas} and \textit{Box Canvas} tasks to the training curriculum. Quantitative and qualitative results are presented in \cref{tab:ablation} and \cref{fig:ablation-task}, respectively. \cref{tab:ablation} clearly show that as more canvas tasks are incorporated, we observe consistent gains in image quality (HPSv3) and control adherence (Control-QA). The qualitative results (\cref{fig:ablation-task}) confirm this: the baseline model fails to follow pose and layout instructions, while the full model successfully handles all multi-control inputs. Additionally, we provide ablations on the impact of the trained branches of MM-DiT\cite{peebles2023scalable} and the convergence behavior of control following in \supp.

%% file: table/ablation.tex
\begin{table}[t]
\centering
\caption{
    \textbf{Ablation study for Multi-Task Canvas Training.} Performance on the Multi-Control Benchmark is evaluated as the \textit{Pose Canvas} and \textit{Box Canvas} tasks are incrementally added to the baseline \textit{Spatial Canvas} model.
}
\label{tab:ablation}
\vspace{-1em}
\resizebox{0.45\textwidth}{!}{%
\begin{tabular}{@{}l ccc ccc ccc@{}}
Model
 & ArcFace$\uparrow$ & VQAScore$\uparrow$ & HPSv3$\uparrow$ & Control-QA$\uparrow$\\
\midrule
Spatial Canvas & \textbf{0.389} & 0.865 & 10.786 & 4.156 \\ 
+ Pose Canvas & 0.371 & 0.874 & 11.440 & 4.188 \\ 
+ Box Canvas & 0.375 & \textbf{0.906}  & \textbf{12.044} & \textbf{4.281} \\ 
\end{tabular}%
}
\end{table}

%% file: sec/5_conclusion.tex
\section{Conclusion}
\label{sec:conclusion}

We introduced \methodname{}, a unified framework for flexible, compositional image generation. Our approach enables a diffusion model to reason jointly over reference subjects, pose signals, and layout constraints by reformulating these heterogeneous controls into a single canvas-conditioned paradigm. Our Multi-Task Canvas training enables \methodname{} to generalize from single-control training samples to complex multi-control scenarios at inference, allowing a single unified model to achieve strong identity preservation, pose fidelity, and structural coherence. This unified canvas formulation establishes a scalable paradigm for multi-modal guidance; while currently bounded by the information density of a single RGB interface, as discussed in the \supp, it establishes a robust foundation for future work to enable even richer forms of visual and semantic control.

\paragraph{Acknowledgements.} We sincerely thank Daniel Cohen-Or and Or Patashnik for carefully reviewing our manuscript and for providing valuable feedback and suggestions throughout this project.

%% file: sec/X_suppl.tex
\clearpage
\setcounter{page}{1}

\maketitlesupplementary

\renewcommand{\thesection}{\Alph{section}}
\renewcommand{\thetable}{\Roman{table}}
\renewcommand{\thefigure}{\Roman{figure}}

\setcounter{section}{0}
\setcounter{table}{0}
\setcounter{figure}{0}

\section{Comparisons with Personalization Methods} \label{sec:pers-comp}

In the main paper, we compare \methodname{} primarily with approaches designed to naturally support the composition task with free form inputs, such as \citep{wu2025qwen, comanici2025gemini, ilkerzgi2025overlay, zhang2025creatidesign}. However, acknowledging that composition and personalization are intersecting tasks, we provide supplementary comparisons with recent zero-shot personalization methods capable of supporting multiple-concept input, specifically UniPortrait~\citep{he2024uniportrait}, FLUX Kontext~\citep{fluxkontext}, UNO~\cite{UNO}, OmniGen2~\citep{ wu2025omnigen2}, DreamO~\cite{DreamO}, and ID-Patch~\cite{ID-Patch}. To ensure a robust evaluation, we extend these comparisons across three distinct benchmarks, detailed below.

\paragraph{4P Composition Benchmark.} Following the 4P Composition setup introduced in the main paper, we provide quantitative comparisons with personalization methods, including \cite{DreamO, wu2025omnigen2, he2024uniportrait} and \cite{fluxkontext} in \cref{tab:supp:4p_composition}. We validate that \methodname{} achieves superior identity preservation (ArcFace), image quality (HPSv3), and text-image alignment (VQAScore) compared to these competing baselines. 
In addition to the quantitative metrics, we provide comparative qualitative examples in Fig. \ref{fig:4p-supp} for a visual assessment. 
These examples illustrate how \methodname{} achieves more consistent identity preservation and realistic multi-subject composition compared to all personalization-based baselines as well as Qwen-Image-Edit~\cite{wu2025qwen}.

\input{fig/training-dynamics}

\input{table/supp-4p}
\input{table/supp-pose}
\input{table/supp_id_obj}

\input{fig/4p-supp}
\input{fig/4p-pose-supp}
\input{fig/id-obj-supp}
\input{fig/pose_2p_supp}
\input{fig/pose_1p_supp}

\paragraph{Pose-Guided 4P Composition Benchmark.} We provide additional quantitative and qualitative comparisons with ID-Patch~\cite{ID-Patch}, a method specifically designed for pose-guided composition with human identities. These results are detailed in Table \ref{tab:supp:pose_comparison}.

To rigorously evaluate this task, we employ a comprehensive set of metrics: ArcFace~\cite{ArcFace} for identity, HPSv3~\cite{ma2025hpsv3} for aesthetic quality, VQAScore~\cite{VQAScore} for semantic alignment, and our proposed Control-QA score (see Sec. \ref{sec:eval-details}). Furthermore, we introduce the PoseAP\textsubscript{0.5} score, which reports the Average Precision (AP\textsubscript{0.5}) for extracted pose keypoints to strictly measure spatial adherence.

As shown in Table \ref{tab:supp:pose_comparison}, Control-QA provides a unified evaluation criterion that jointly considers pose accuracy and identity preservation (measured by ArcFace similarity), under which our method achieves the highest score. Qualitative results in Fig. \ref{fig:4p-pose-supp} further demonstrate that although ID-Patch~\cite{ID-Patch}, through its integration with ControlNet~\cite{zhang2023adding}, can effectively reproduce target poses—resulting in high PoseAP scores—it often fails to maintain the correct number of subjects and consistent identities. In contrast, \methodname{} achieves a more balanced trade-off between pose fidelity and identity preservation. Additional qualitative examples of pose-guided composition in single-person (1P) and two-person (2P) scenarios are presented in Fig. \ref{fig:pose-1p-supp} and Fig. \ref{fig:pose-2p-supp}, respectively.

\paragraph{ID-Object Interaction Benchmark.} To demonstrate the generalizability of our approach beyond human subjects, and to evaluate performance in scenarios involving natural interactions between subjects, we extend our evaluations to the ID-Object Interaction benchmark. To construct this benchmark, we pair human identities from the FFHQ-in-the-Wild~\cite{StyleGAN} dataset with object references from the DreamBooth~\cite{ruiz2023dreambooth} dataset to create challenging ID–Object pairs.

We quantitatively compare our method against a wide range of baselines, including \cite{UNO, DreamO, wu2025omnigen2, ilkerzgi2025overlay, fluxkontext, comanici2025gemini}, as well as our main baseline, Qwen-Image-Edit~\cite{wu2025qwen}. Corresponding quantitative results are provided in Table \ref{tab:id_obj_composition}.
Our \methodname{} achieves the highest identity and object preservation, as well as the strongest overall control following, as indicated by ArcFace, DINOv2~\cite{oquabdinov2}, and Control-QA metrics, respectively. To further assess interaction fidelity, we provide qualitative comparisons in \cref{fig:id-obj-supp}. \methodname{} produces coherent compositions that faithfully preserve both human identity and object fidelity, maintaining correct proportions and natural interactions between them, whereas existing baselines often fail to achieve realistic integration of the two.

\section{Supplementary Ablations}

In addition to the ablation studies in the main paper, we provide a deeper analysis of the training dynamics of \methodname{}. Specifically, we examine the convergence behavior under multi-task learning and empirically validate our selection of trainable blocks.

\paragraph{Convergence Behavior of \methodname{}.} We tracked the model’s performance across different training iterations (\cref{fig:supp:training-dynamics}). The Control-QA curve shows steady improvement in the early stages, with rapid gains up to 50K iterations, where convergence is largely achieved. During this phase, the model progressively strengthens control adherence. Although key metrics plateau beyond 50K, we continue training up to 200K iterations to refine local details and improve robustness. All subsequent ablation studies use this 200K-iteration model as the default checkpoint.

\paragraph{Ablations of Trainable Blocks.} We investigate the impact of different architectural choices for LoRA optimization. In our default configuration, we train modulation and attention layers within the text and image attention branches, while keeping feed-forward layers frozen. Table \ref{tab:supp-ablation} quantifies the impact of including or excluding these components on the 4P Composition benchmark. Two key findings emerge from this analysis. First, effective identity preservation requires the joint training of both the text and image branches; omitting either leads to a drop in identity fidelity. Second, training the feed-forward layers negatively impacts the model's generalization; we observe a deterioration in both visual quality and prompt alignment when these layers are unfrozen. Based on these results, our final model excludes feed-forward layers from the optimization process. Finally, we evaluate the contribution of the task indicator prompt ($c$). As reported in \cref{tab:supp-ablation}, removing this indicator leads to a degradation in performance across all metrics. This confirms that explicitly signaling the control type is crucial for the model to resolve ambiguity and effectively switch between different compositional reasoning modes. We provide qualitative ablations on the task indicator in Fig. \ref{fig:task-indicator}

\input{table/supp_ablation}

\input{fig/task-inticator}

\input{fig/bg-guided}

\section{Limitations}
While \methodname{} provides an intuitive interface for compositional image generation with multimodal inputs, enabling combined controls in a single inference pass, the ``visual canvas'' format has inherent constraints. Although this format offers significant advantages in usability and flexibility, it is strictly bound by the available pixel space. As demonstrated in Fig. \ref{fig:4p-bench} and \ref{fig:pose-bench}, \methodname{} successfully handles occluding entities up to 4P composition, outperforming baseline approaches. However, relying on a single RGB canvas implicitly limits the number of concepts that can be interpreted simultaneously; as the number of concepts increases, the canvas becomes crowded and harder to interpret. To resolve this, future work could explore layered controls, such as designing the input canvas with an additional alpha channel (RGBA).

\section{Additional Applications}
\methodname{} is also capable of background-aware composition. We provide qualitative examples of this capability in \cref{fig:bg-guided}. \methodname{} can inject humans or objects into a scene through reference image pasting or bounding box annotation, with the inserted elements naturally interacting with the background.

\section{User Study} \label{sec:user-study}

We validate the effectiveness of \methodname{} on the Multi-Control Composition task through human evaluation. To ensure a fair and accurate assessment, we conduct two separate user studies aimed at evaluating the condition-following behavior of \methodname{} against competing methods. Given the cognitive difficulty of assessing three simultaneous conditions (pose, identity, and box layout) at once, we decouple the input controls into two distinct pairwise comparisons: Pose + Identity'' and Pose + Box Layout''.

For each combination, we perform a separate study with unique participants. In total, we collected responses from 30 anonymous participants for 30 examples per study, conducted via the \textit{Prolific} platform. The specific setups are detailed below:

\begin{itemize} 
    \item \textbf{Control Following (Pose + Box Layout''):} This study focuses on the structural capabilities of the model. For each question, users are shown an input pose reference and a box layout. Then they are presented with generated samples and asked to select which generation better adheres to the input controls. We utilize an A/B testing setup in which users select their preferred output. The instructions provided to the participants are shown in Fig. \ref{fig:user-study-1-instr}, and a sample question is provided in Fig. \ref{fig:user-study-1-question}. 
    \item \textbf{Identity Preservation (Pose + Identity''):} To evaluate identity fidelity under spatial constraints, this study focuses on how well the subject's identity is preserved while applying a specific pose. Users are instructed to prioritize identity preservation in their assessment while verifying that the pose is applied. Similar to the previous study, we use an A/B setup. User instructions are provided in Fig. \ref{fig:user-study-2-instr}, with a sample question in Fig. \ref{fig:user-study-2-question}. 
\end{itemize}

We report the win rates against competing methods in Table \ref{tab:user_study} for both the Control Following'' and Identity Preservation'' evaluations. Consistent with the quantitative analyses in the Multi-Control Composition Benchmark, we compare our method against \cite{wu2025qwen} and \cite{comanici2025gemini}.

Our results indicate that while \methodname{} outperforms both baselines overall, there is a distinct trade-off among competitors: \cite{comanici2025gemini} performs stronger on identity preservation, whereas \cite{wu2025qwen} performs better on control following. This alignment between human preference and our reported metrics serves as a strong validation for our proposed Control-QA score, confirming its success as a unified metric for evaluating multiple control inputs.

\input{table/supp_user_study}

\section{Benchmark Details}\label{sec:eval-details}
\subsection{Evaluation Metrics}
For all evaluations, we employ a unified setup focusing on identity preservation, visual quality, prompt alignment, and control adherence. We detail the specific metrics below.

\paragraph{ArcFace \& DINOv2.} We use ArcFace~\cite{ArcFace} to quantitatively evaluate identity preservation across all benchmarks involving human subjects (i.e., 4P Composition, Pose-Guided 4P Composition, Multi-Control Composition, and ID-Object Interaction). We sample input identities from the FFHQ-in-the-wild~\cite{StyleGAN} dataset and compute the ArcFace similarity score between the generated image and the masked identities in the corresponding composition. For object consistency in the ID-Object Interaction Benchmark, we utilize DINOv2~\cite{oquabdinov2} in a similar manner to calculate similarity scores.

\paragraph{HPSv3 and VQAScore.} To evaluate visual quality and adherence to the input prompt, we use the Human Preference Score v2 (HPSv3)~\cite{ma2025hpsv3}. We compute this metric using the original generation prompt. Although closed-source models like \cite{comanici2025gemini} may employ internal prompt rewriting, we ignore such implicit augmentations to ensure a fair comparison based on the user-provided input. Additionally, we utilize VQAScore~\cite{VQAScore} to further assess prompt alignment across our experiments.

\paragraph{Control-QA.} Given the variety of control settings in \methodname{} (e.g., pose, spatial layout, layout boxes), we establish a unified evaluation framework using an LLM-based scoring system. We employ GPT-4o~\cite{ChatGPT4o} as a multimodal expert to rate the generated compositions against the provided control images. The system prompts used for the 4P Composition, Pose-Guided 4P Composition, Layout-Guided Composition, and Multi-Control Composition benchmarks are provided in Tables \ref{tab:system_prompt_small}, \ref{tab:system_prompt_pose}, \ref{tab:system_prompt_spatial}, and \ref{tab:system_prompt_joint}, respectively. Note that additional quantitative evaluations for pose adherence (PoseAP) are discussed in Sec. \ref{sec:pers-comp}, and human user studies are detailed in Sec. \ref{sec:user-study}.

\subsection{Evaluation Benchmarks}

We employ an automated pipeline to construct the input canvases for all benchmarks. Below, we detail the construction process for each specific task.

\paragraph{4P Composition Benchmark.} To construct the canvases for the 4P Composition benchmark, we randomly sample four human identities from the FFHQ-in-the-wild dataset~\cite{StyleGAN}. To determine a natural spatial arrangement for these individuals, we employ a two-step process. First, we generate a synthetic ``prior image'' using FLUX.1-Dev~\cite{FLUX} based on the target prompt. Second, we detect the human instances within this prior image to obtain realistic bounding boxes. Finally, we construct the input canvas by placing the segmented FFHQ identities into these extracted positions.

\paragraph{Pose-Guided 4P Composition Benchmark.} Building upon the 4P Composition setup, we incorporate structural control into the pipeline. We utilize the same FLUX.1-Dev~\cite{FLUX} prior images generated for the 4P task, but instead of just extracting bounding boxes, we utilize our internal pose estimation model to extract the target poses. We then construct the input canvas by placing these target poses alongside the reference identities.

\paragraph{Layout-Guided Composition Benchmark.} As this benchmark focuses on named entity composition based on a layout rather than human identity, we utilize the test set of the CreatiDesign~\cite{zhang2025creatidesign} dataset. Since our canvas format utilizes text overlaid directly on the image (rather than regional prompting), we filter the test set to select samples compatible with this modality. It is worth noting that the CreatiDesign dataset places a strong emphasis on text rendering capabilities, as demonstrated in our qualitative comparisons (see Fig. \ref{fig:creatidesign}).

\paragraph{Multi-Control Composition Benchmark.} For this complex setting, we leverage the text prompts and named entity annotations from the CreatiDesign~\cite{zhang2025creatidesign} test set, specifically filtering for samples that involve human subjects. To obtain a valid target pose that aligns with these prompts, we generate a synthetic prior image using our baseline model, Qwen-Image-Edit~\cite{wu2025qwen}. Crucially, we do not utilize the pixel data of this prior image as a direct input; instead, we use it strictly to extract the target skeletal pose. We then construct the final input canvas by combining this extracted pose, a sampled reference identity, and the named entity annotations (for text rendering) from the original CreatiDesign sample. This setup simultaneously evaluates identity preservation, pose adherence, and text rendering, as highlighted in Fig. \ref{fig:mixed-control} and \ref{fig:ablation-task}.

\subsection{Dataset Details}To train \methodname{}, we utilize an internal cross-frame dataset augmented with the CreatiDesign~\cite{zhang2025creatidesign} dataset. Our internal dataset comprises $\sim$6M human-centric training images, constituting $\sim$1M scenes with cross-frame samples. Due to legal constraints, we cannot open-source this internal dataset; however, a similar multi-frame dataset can be constructed from public open-source video datasets. From these 1M scenes, we use an internal instance segmentation model to extract human segments for constructing the input canvases, while treating the remaining image areas as the background. Similarly, we extract poses from the target frames using an internal pose estimation model. Since \methodname{} is built upon this human-centric data, we construct the human boxes in the ``Box Canvas'' using these extracted segments. To enable the model to be capable with a variety of objects, we include the CreatiDesign~\cite{zhang2025creatidesign} dataset, which introduces named annotations into our training set along with text-rendering focused samples.

\input{fig/user-study-1}

\input{table/system-prompts}

%% file: fig/training-dynamics.tex
\begin{figure}[!t]
    \centering
    \includegraphics[width=\linewidth]{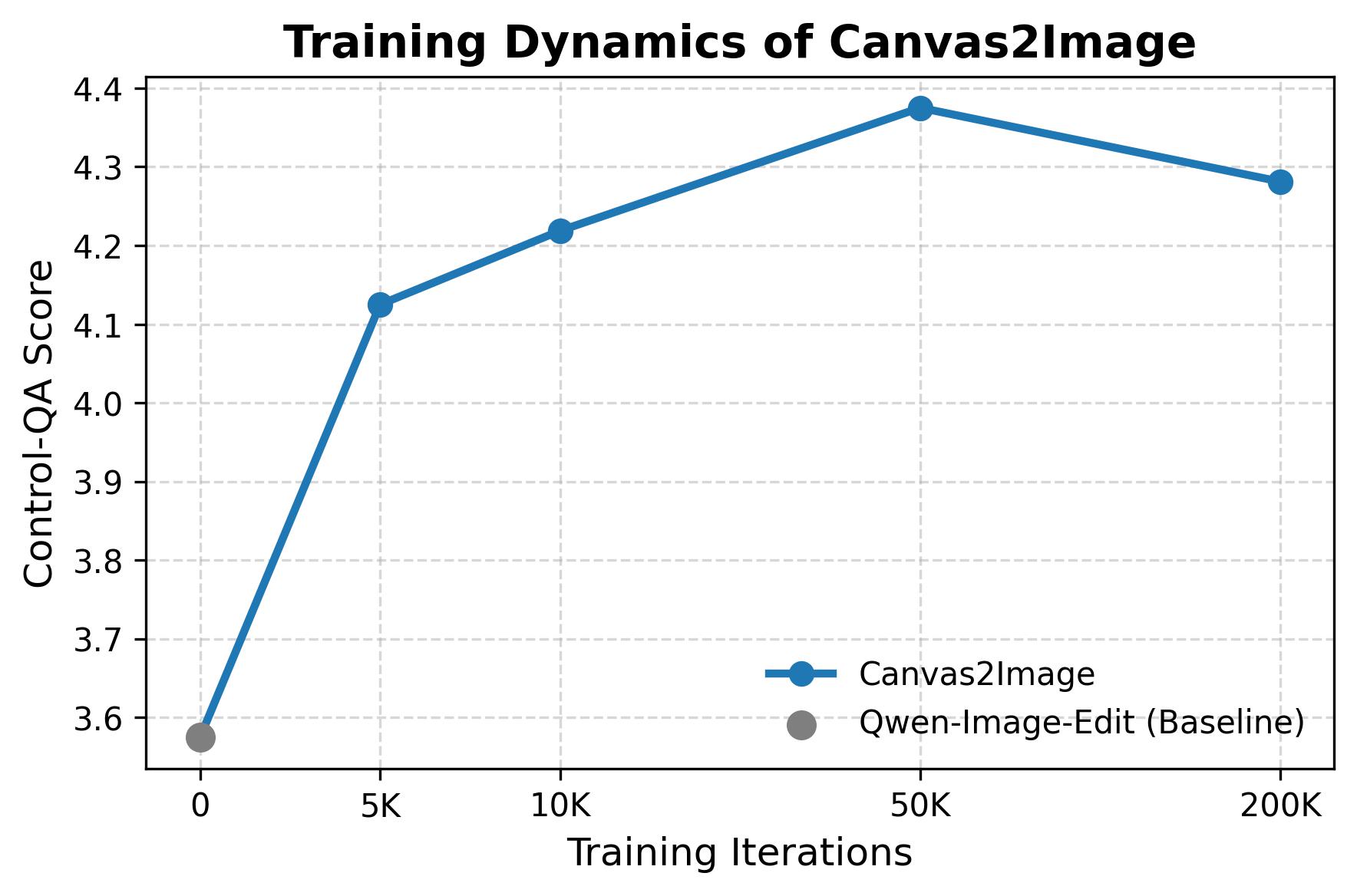}
\caption{\textbf{Training Dynamics for \methodname{}.} The Control-QA score steadily improves during early training and converges around 50K iterations, indicating that the model effectively learns consistent control and composition. We further train up to 200K iterations to refine local details and enhance robustness in generation quality.}
    \label{fig:supp:training-dynamics}
\end{figure}

%% file: table/supp-4p.tex
\begin{table}[t]
\centering
\caption{\textbf{Quantitative Comparison Including Personalization Baselines on the 4P Composition Benchmark.} \textbf{Bold} values denote the best performance for each metric, highlighting the superior overall performance of our method across all categories.}
\label{tab:supp:4p_composition}
\resizebox{\columnwidth}{!}{%
\begin{tabular}{lcccc}
\toprule
& \textbf{ArcFace $\uparrow$} & \textbf{HPSv3 $\uparrow$} & \textbf{VQAScore $\uparrow$} & \textbf{Control-QA $\uparrow$} \\
\midrule
DreamO \cite{DreamO} & 0.2049 & 12.4210 & 0.7782 & 1.4062 \\
OmniGen2 \cite{wu2025omnigen2} & 0.0859 & 12.9873 & 0.8051 & 1.9688 \\
ID-Patch \cite{ID-Patch} & 0.0824 & 7.1262 & 0.7846 & 1.0938 \\
UniPortrait \cite{he2024uniportrait} & 0.3088 & 12.4011 & 0.7860 & 2.5000 \\
Overlay Kontext \cite{ilkerzgi2025overlay} & 0.1709 & 12.6932 & 0.8792 & 2.0000 \\
Flux Kontext \cite{fluxkontext} & 0.2168 & 12.7684 & 0.8687 & 2.2188 \\
UNO \cite{UNO} & 0.0769 & 12.1558 & 0.8402 & 1.5000 \\
Nano Banana \cite{comanici2025gemini} & 0.4335 & 10.3857 & 0.8260 & 3.8750 \\
Qwen Image Edit \cite{wu2025qwen} & 0.2580 & 13.1355 & 0.8974 & 3.6875 \\
\textbf{Ours} & \textbf{0.5915} & \textbf{13.2295} & \textbf{0.9002} & \textbf{4.0000} \\
\bottomrule
\end{tabular}%
}
\end{table}

%% file: table/supp-pose.tex
\begin{table}[t]
\centering
\caption{\textbf{Quantitative Comparisons on the Pose-Guided 4P Composition Benchmark.} The Control-QA score provides a unified criterion that accounts for both pose accuracy and identity preservation, where our method achieves the highest performance among all baselines.}
\label{tab:supp:pose_comparison}
\resizebox{\columnwidth}{!}{%
\begin{tabular}{lccccc}
\toprule
\textbf{Pose} & \textbf{ArcFace} $\uparrow$ & \textbf{HPSv3} $\uparrow$ & \textbf{VQAScore} $\uparrow$ & \textbf{Control-QA} $\uparrow$ & \textbf{PoseAP\textsubscript{0.5}} $\uparrow$ \\
\midrule
ID-Patch \cite{ID-Patch} & \underline{0.2854} & 11.9714 & \underline{0.8955} & \underline{4.1250} & \textbf{75.0814 }\\
Nano Banana \cite{comanici2025gemini} & 0.2623 & 9.9727 & 0.8609 & 3.4375 & 64.1704 \\
Qwen-Image-Edit \cite{wu2025qwen} & 0.1534 & \textbf{12.9397} & 0.8897 & 4.0312 & 67.2734 \\
\textbf{Ours} & \textbf{0.3001} & \underline{12.8989} & \textbf{0.8971} & \textbf{4.4688} & \underline{70.1670} \\
\bottomrule
\end{tabular}
}
\end{table}

%% file: table/supp_id_obj.tex
\begin{table}
\centering
\caption{\textbf{Quantitative Results on the ID-Object Composition Benchmark.} We compare our method with several baselines across five different metrics. \textbf{Bold} values indicate the best performance in each column. DINOv2 measures object preservation. Our \methodname{} achieves the highest identity (ArcFace) and object (DINOv2~\cite{oquabdinov2} preservation as well as the highest control following (Control-QA). }
\label{tab:id_obj_composition}
\resizebox{\columnwidth}{!}{%
\begin{tabular}{lccccc}
\toprule & \textbf{ArcFace $\uparrow$} & \textbf{HPSv3 $\uparrow$} & \textbf{VQAScore $\uparrow$} & \textbf{Control-QA $\uparrow$} & \textbf{DINOv2 $\uparrow$} \\
\midrule
UNO \cite{UNO} & 0.0718 & 8.6718 & 0.8712 & 2.5000 & 0.2164 \\
DreamO \cite{DreamO} & 0.4028 & 9.0394 & 0.8714 & 3.9688 & 0.3111 \\
OmniGen2 \cite{wu2025omnigen2} & 0.1004 & 10.2854 & \textbf{0.9266} & 4.4062 & 0.3099 \\
Overlay Kontext \cite{ilkerzgi2025overlay} & 0.1024 & 8.6132 & 0.8539 & 3.2812 & 0.2703 \\
Flux Kontext \cite{fluxkontext} & 0.1805 & 9.2179 & 0.8914 & 3.1562 & 0.2818 \\
Qwen-Image-Edit \cite{wu2025qwen} & 0.3454 & \textbf{10.3703} & 0.9045 & 4.4062 & 0.2867 \\
\textbf{Ours} & \textbf{0.5506} & 9.7868 & 0.9137 & \textbf{4.8750} & \textbf{0.3298} \\
\bottomrule
\end{tabular}%
}
\end{table}

%% file: fig/4p-supp.tex
\begin{figure*}[!t]
    \centering
    \includegraphics[width=\linewidth]{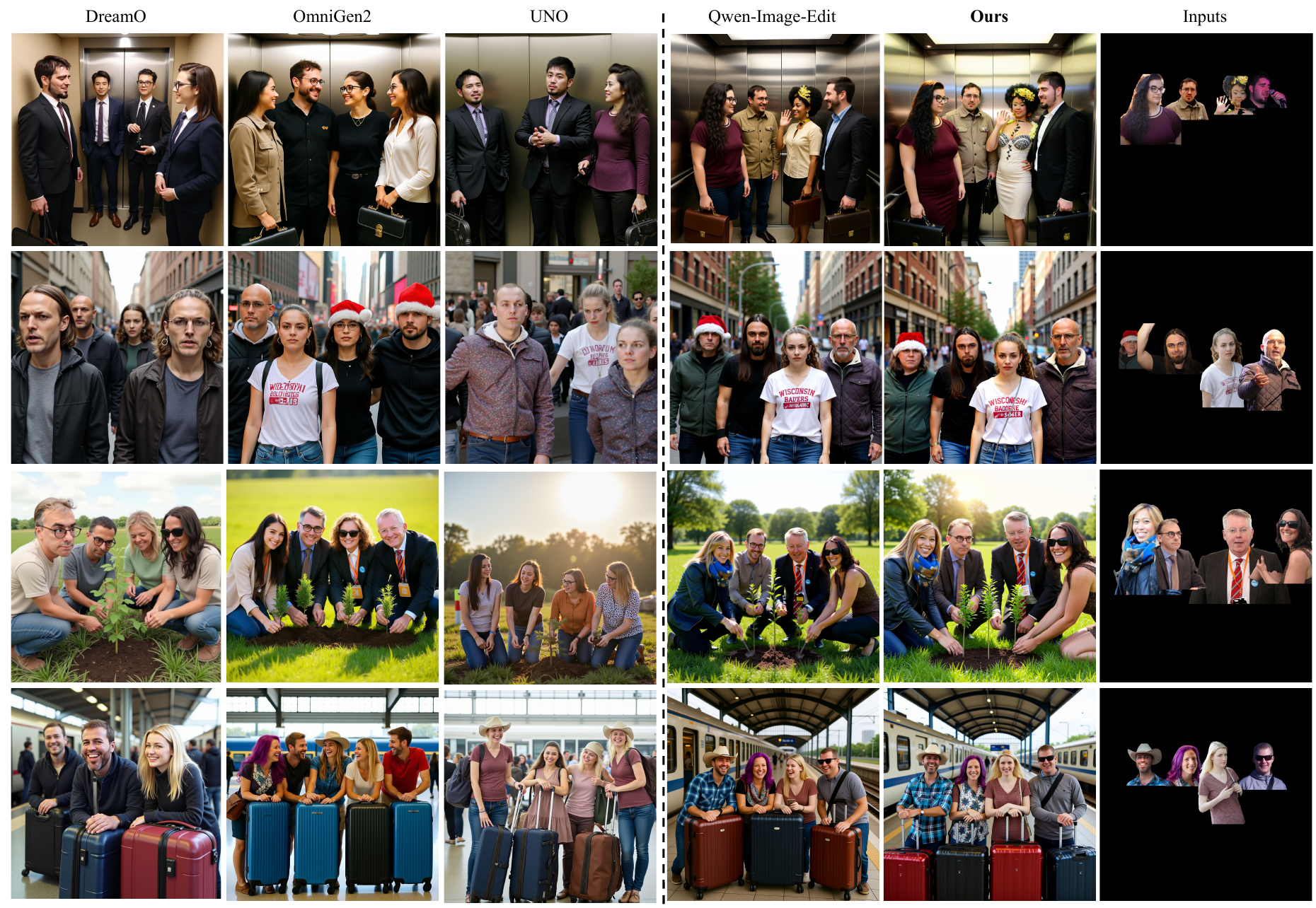}
\caption{\textbf{Supplementary Qualitative Comparisons on the 4P Composition Benchmark with Personalization Approaches.} Both Qwen-Image-Edit and our \methodname{} significantly outperform the state-of-the-art multi-subject personalization baselines DreamO~\cite{DreamO}, OmniGen2~\cite{wu2025omnigen2}, and UNO~\cite{UNO} in terms of identity preservation. Compared to Qwen-Image-Edit, our method demonstrates further improvements in identity fidelity, particularly for the rightmost man in the $1^{st}$ row, the leftmost woman in the $2^{nd}$ row, and the second man in the $3^{rd}$ row.}
    \label{fig:4p-supp}
\end{figure*}

%% file: fig/4p-pose-supp.tex
\begin{figure*}[!t]
    \centering
    \includegraphics[width=\linewidth]{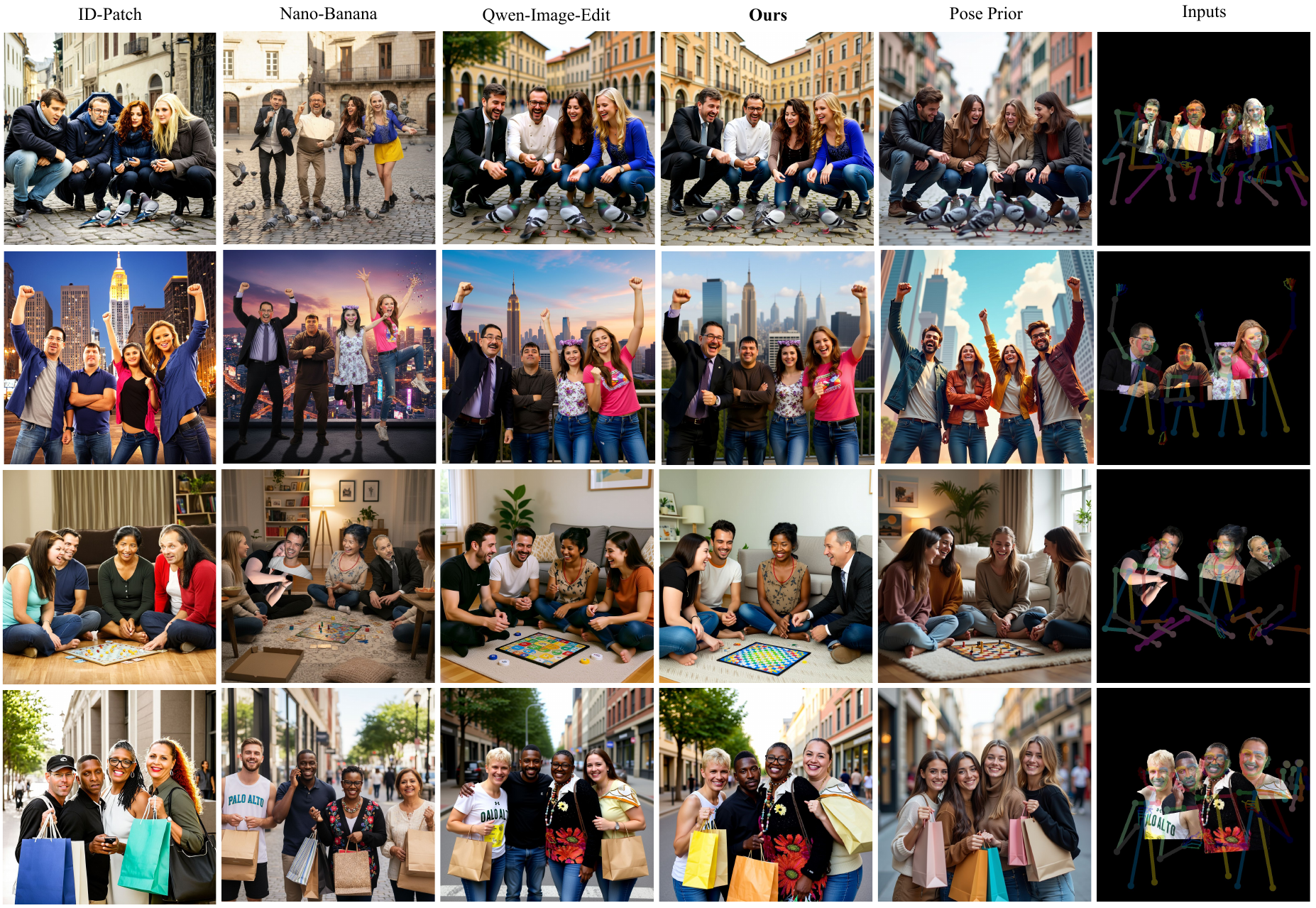}
    \caption{\textbf{Supplementary Qualitative Comparisons on the Pose-Overlaid 4P Composition Benchmark.} We additionally include the relevant state-of-the-art personalization baseline, ID-Patch~\cite{ID-Patch}, in our comparison. While ID-Patch follows poses to some extent, it performs significantly worse in identity preservation and image quality. In contrast, image editing baselines fail to accurately follow the target pose. Our \methodname{} achieves both strong identity preservation and precise pose alignment.}
    \label{fig:4p-pose-supp}
\end{figure*}

%% file: fig/id-obj-supp.tex
\begin{figure*}[!t]
    \centering
    \includegraphics[width=\linewidth]{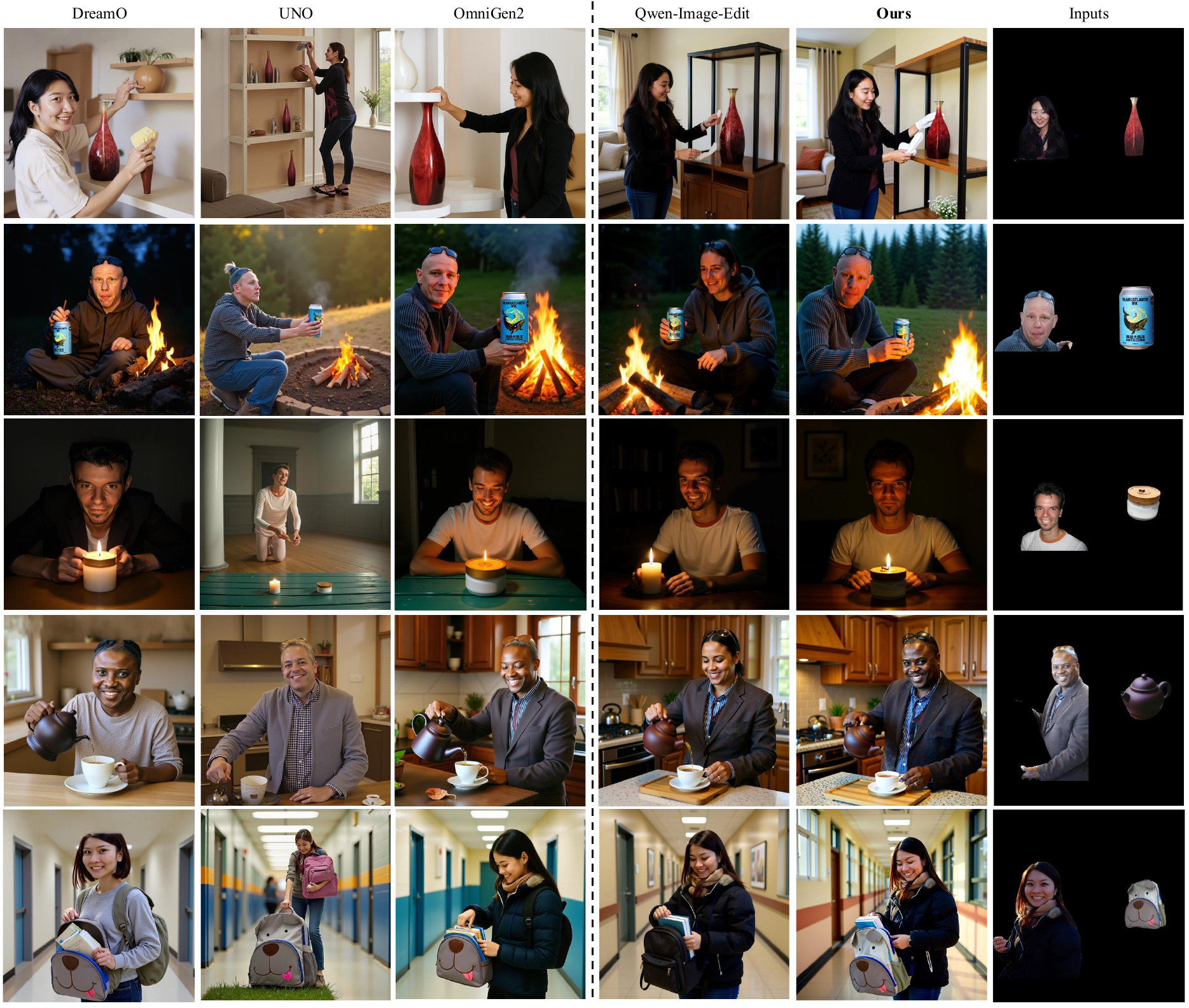}
\caption{\textbf{Qualitative Results on the ID–Object Composition Benchmark.} Our \methodname{} generates coherent compositions that faithfully preserve both human identity and object fidelity, maintaining correct proportions and natural interactions between them. In contrast, existing baselines often fail to achieve realistic integration between the human and the object. For instance, the baseline Qwen-Image-Edit~\cite{wu2025qwen} fails to preserve both identity and object consistency, as illustrated in these examples.}
    \label{fig:id-obj-supp}
\end{figure*}

%% file: fig/pose_2p_supp.tex
\begin{figure*}[!t]
    \centering
    \includegraphics[width=\linewidth]{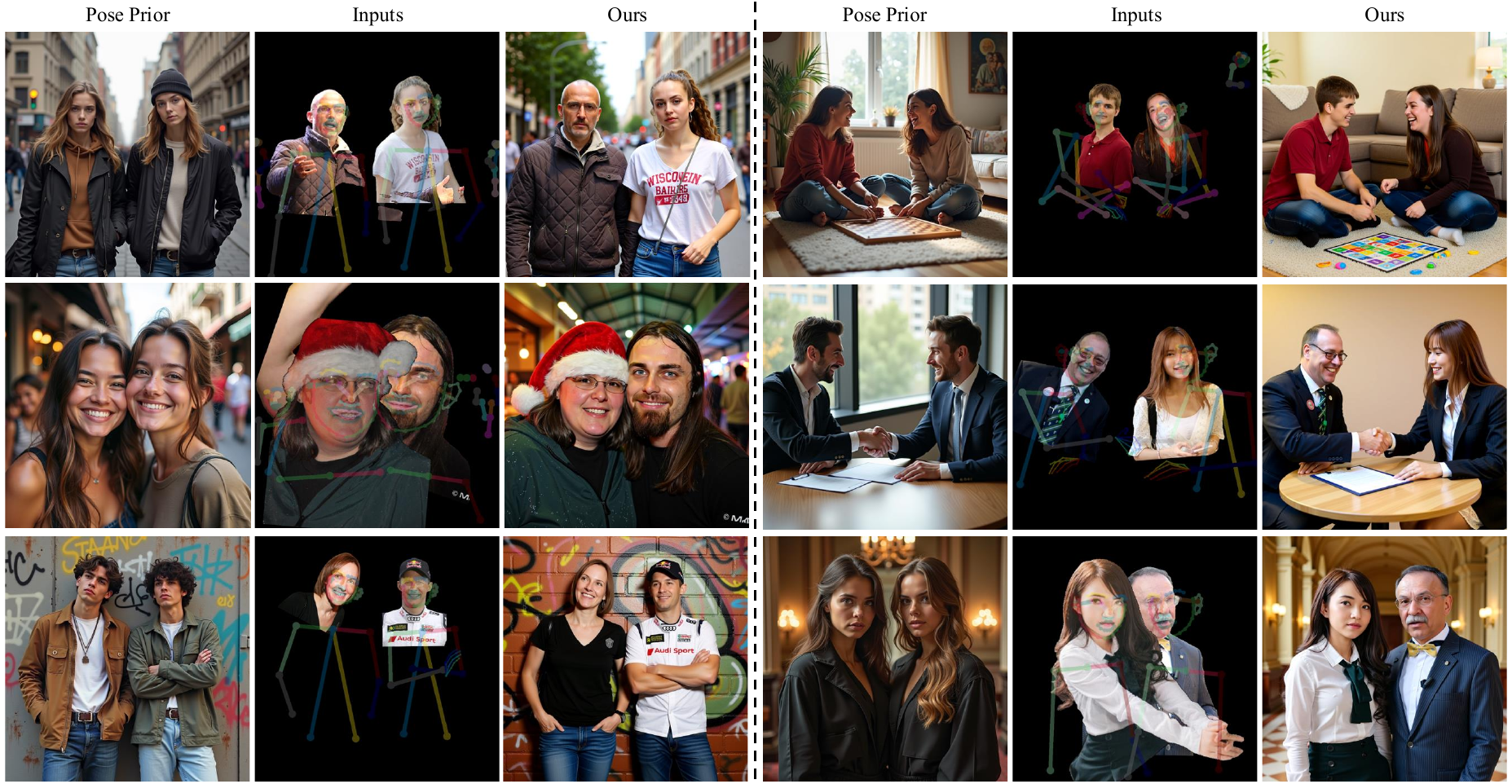}
    \caption{\textbf{Qualitative Results on Pose-Overlaid 2P Composition.} Under the \textit{Pose Canvas} setup, our \methodname{} achieves superior identity preservation and accurate pose alignment. Notably, \methodname{} closely follows the target poses defined by the prior generated from FLUX-Dev~\cite{FLUX} (``Pose Prior'' column), while producing coherent and high-quality images.}
    \label{fig:pose-2p-supp}
\end{figure*}

%% file: fig/pose_1p_supp.tex
\begin{figure*}[!t]
    \centering
    \includegraphics[width=\linewidth]{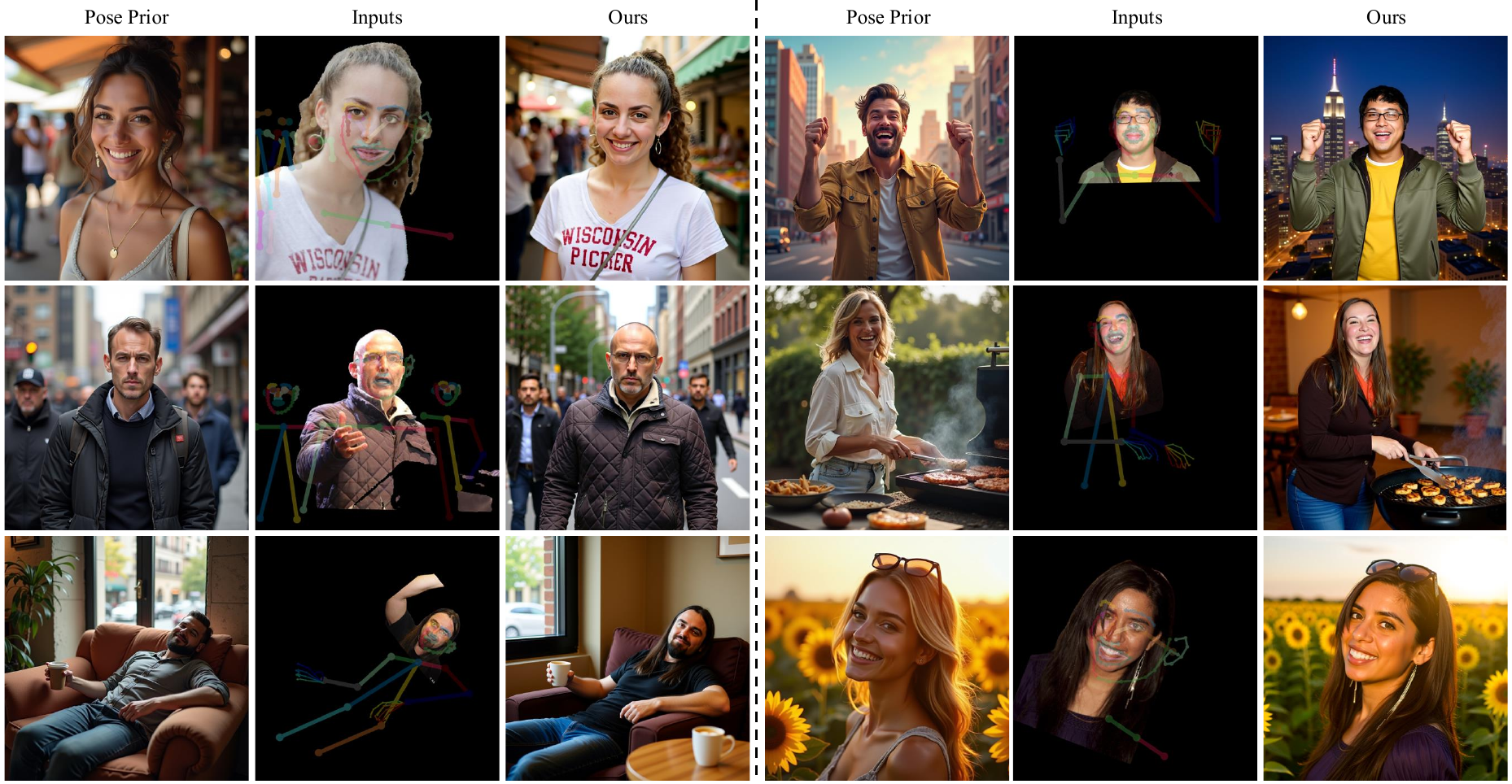}
    \caption{\textbf{Qualitative Results on Pose-Overlaid 1P Composition.}
    Under the \textit{Pose Canvas} setup, our \methodname{} again achieves superior identity preservation and accurate pose alignment. 
    }
    \label{fig:pose-1p-supp}
\end{figure*}

%% file: table/supp_ablation.tex
\begin{table}[t]
\centering
\caption{\textbf{Ablations on Model Architecture.} We conduct ablations on the fine-tuned layers using the 4P Composition Benchmark. Our default configuration, which fine-tunes both the text and image branches while excluding the feed-forward layer, with task indicators included, achieves the highest overall performance.}
\label{tab:supp-ablation}
\resizebox{0.45\textwidth}{!}{
\begin{tabular}{@{}l ccc ccc ccc@{}}
Model
 & ArcFace$\uparrow$ & HPSv3$\uparrow$ & VQAScore$\uparrow$ \\
\midrule
Qwen-Image-Edit & 0.2580 & 13.1355 & 0.8974  \\
Ours w/o Text Branch & 0.4917 & 11.6523 & 0.8297 \\
Ours w/o Image Branch & 0.4687 & 12.7077 & 0.8880 & \\
Ours w/ Feed-Forward & 0.5603 & 12.4846 & 0.8577 \\
Ours w/o Task Indicator & 0.5217 & 12.6046 & 0.8555 \\
\hline
\textbf{Ours} & \textbf{0.5915} & \textbf{13.2295} & \textbf{0.9002}\\ 
\end{tabular}
}
\end{table}

%% file: fig/task-inticator.tex
\begin{figure}
    \centering
    \includegraphics[width=\linewidth]{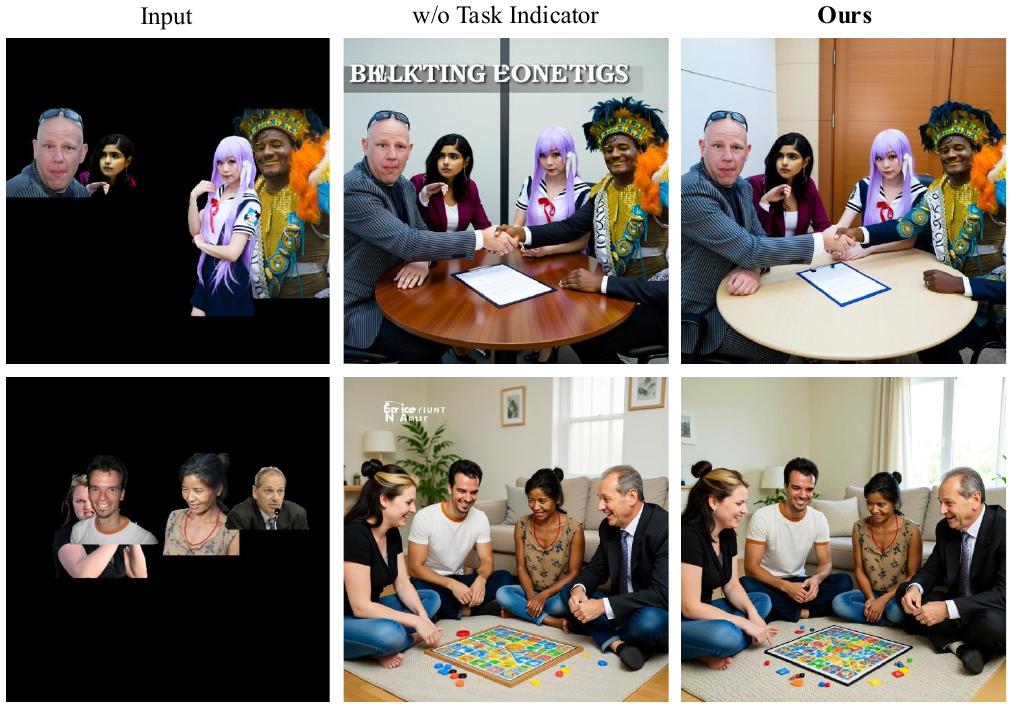}
    \caption{\textbf{Qualitative Ablations on the Task Indicator.} We visualize the impact of removing the task indicator prompt ($c$) in training. Without this explicit signal, the model suffers from task mix-up, where the 4P Composition (Spatial Canvas) is impacted by the Box Canvas task. This results in unwanted text artifacts appearing in the background, as the model incorrectly transfers the text-rendering behavior required only in box-canvas settings to a spatial composition benchmark that does not require text rendering.}
    \label{fig:task-indicator}
\end{figure}

%% file: fig/bg-guided.tex
\begin{figure*}[!t]
    \centering
    \includegraphics[width=\linewidth]{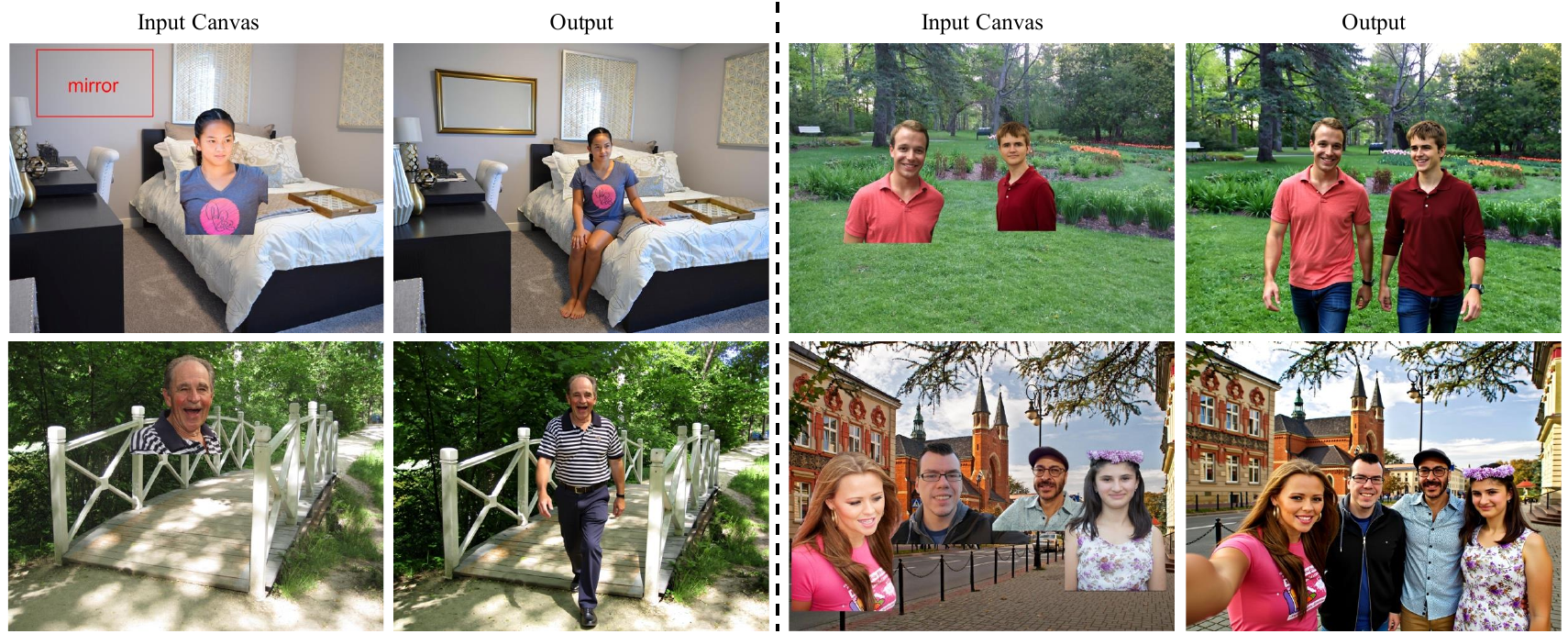}
\caption{\textbf{Background-Aware Composition with \methodname{}.} Given a background image, \methodname{} seamlessly integrates humans or objects into the scene through reference image pasting or bounding box annotations, producing natural spatial alignment and consistent lighting with the surrounding environment.}
    \label{fig:bg-guided}
\end{figure*}

%% file: table/supp_user_study.tex
\begin{table}[h]
    \centering
\caption{\textbf{User Study Results.} The win rate represents the percentage of cases where users preferred our results over baseline methods. \methodname{} is significantly preferred in both control following and identity preservation compared to strong baselines.}
    \label{tab:user_study}
    \resizebox{\linewidth}{!}{%
    \begin{tabular}{lcc}
        \toprule
         & \textbf{Control Following} & \textbf{Identity Preservation} \\
        \midrule
        Ours vs. Qwen-Image-Edit \cite{wu2025qwen} & \textbf{67.3}\% & \textbf{77.3}\% \\
        Ours vs. Nano Banana \cite{comanici2025gemini} & \textbf{78.9}\% & \textbf{73.8}\% \\
        \bottomrule
    \end{tabular}
    }
\end{table}

%% file: fig/user-study-1.tex
\begin{figure}
    \centering
    \includegraphics[width=\linewidth]{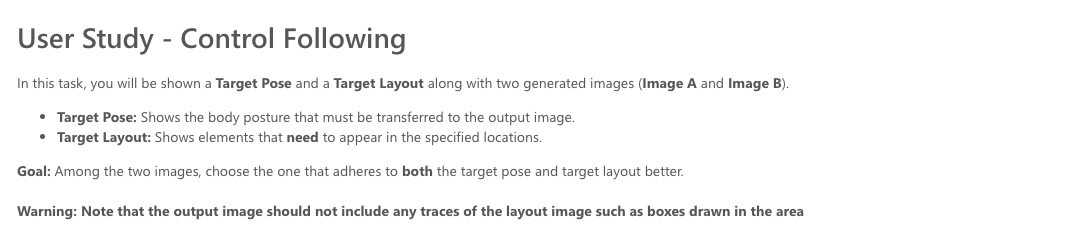}
    \caption{\textbf{User Instructions for User Study ``Control Following''.}}
    \label{fig:user-study-1-instr}
\end{figure}

\begin{figure}
    \centering
    \includegraphics[width=\linewidth]{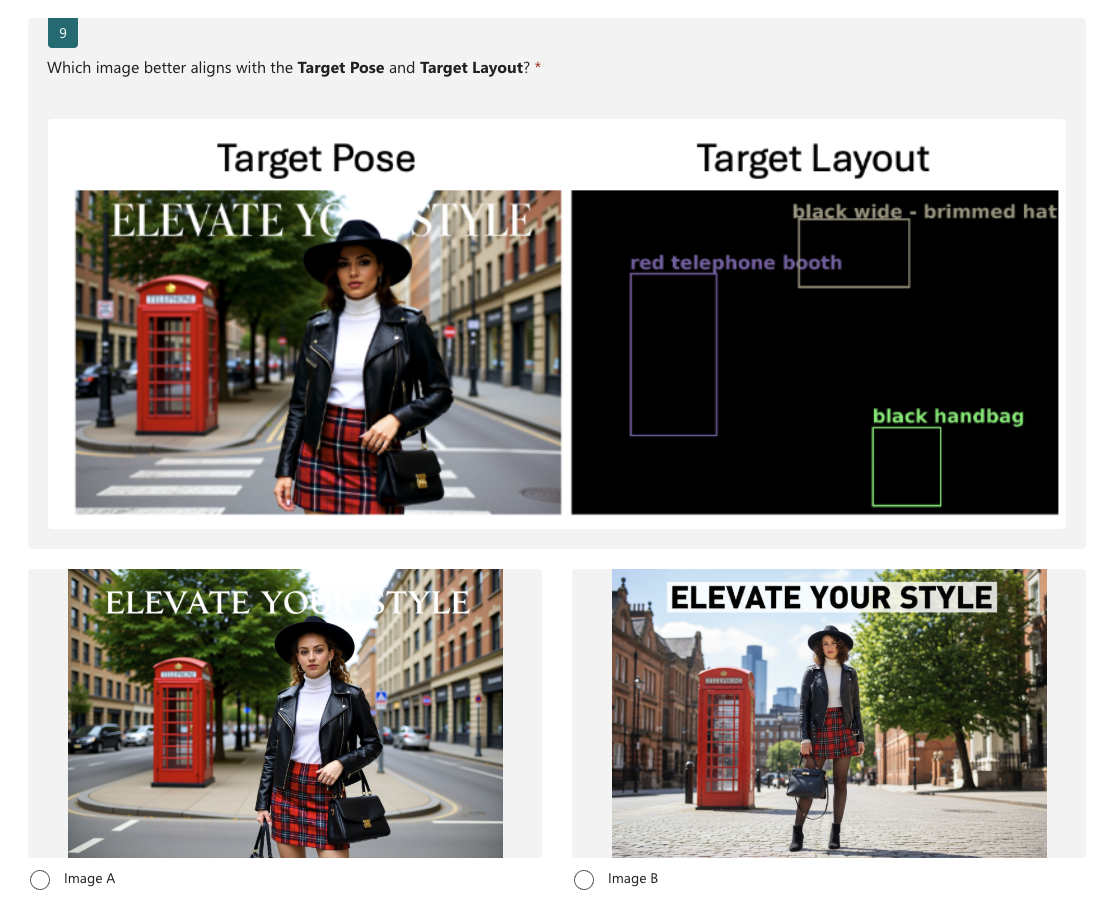}
    \caption{\textbf{Sample Question for User Study ``Control Following''.}}
    \label{fig:user-study-1-question}
\end{figure}

\begin{figure}
    \centering
    \includegraphics[width=\linewidth]{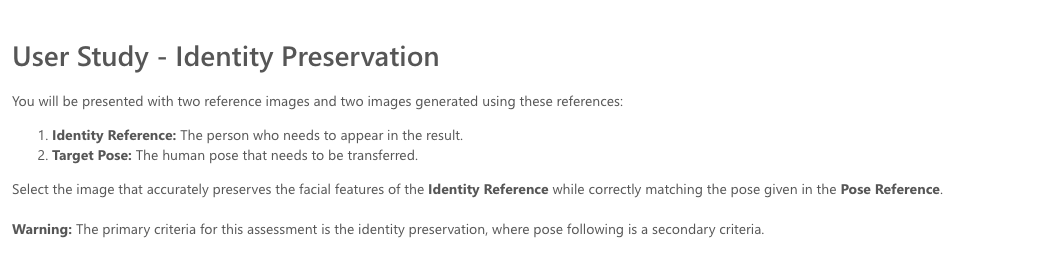}
    \caption{\textbf{User Instructions for User Study ``Identity Preservation''.}}
    \label{fig:user-study-2-instr}
\end{figure}

\begin{figure}
    \centering
    \includegraphics[width=\linewidth]{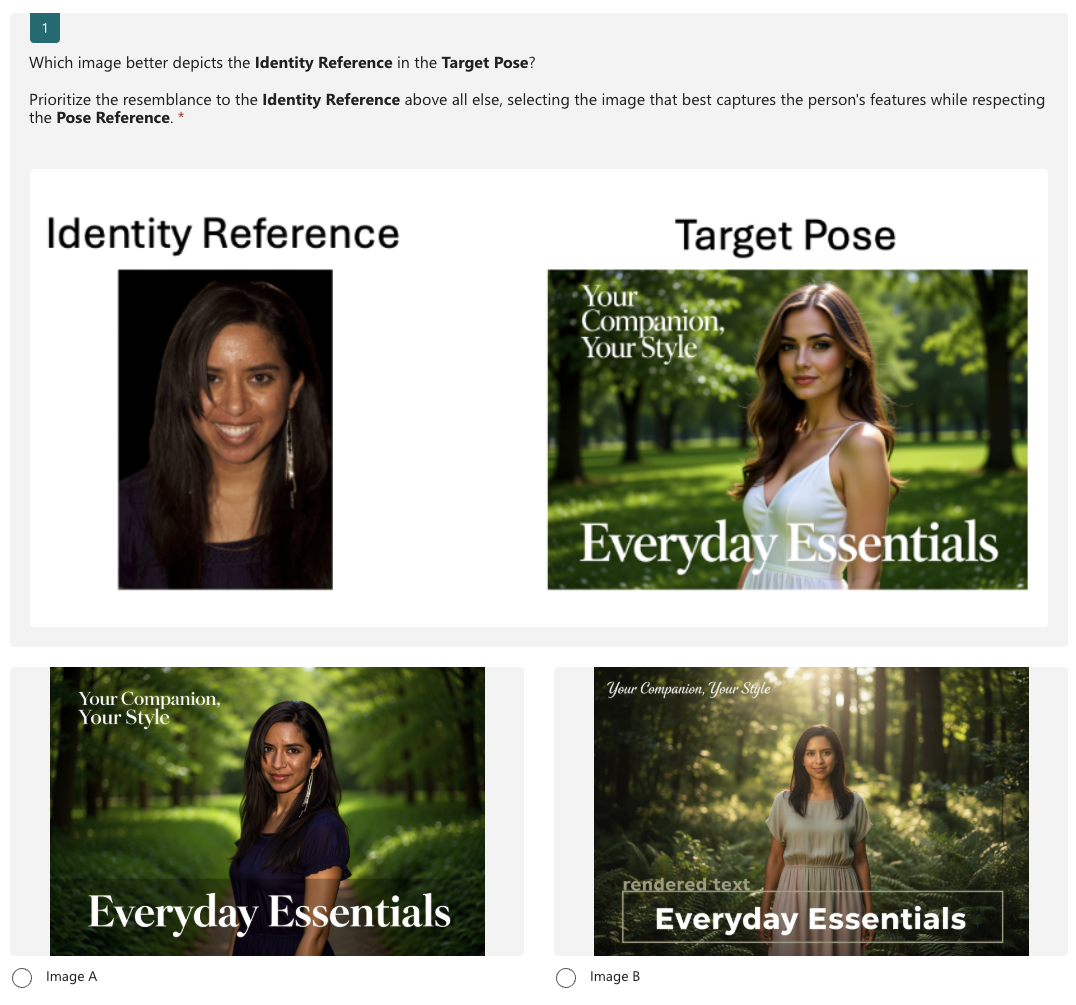}
    \caption{\textbf{Sample Question for User Study ``Identity Preservation''.}}
    \label{fig:user-study-2-question}
\end{figure}

%% file: table/system-prompts.tex
\begin{table*}[t]
\centering
\caption{\textbf{Compositional Fidelity Evaluation Protocol (System Prompt).} This protocol was provided to the human evaluators and served as the instruction set for the LLM-based scoring system.}
\label{tab:system_prompt_small}

\begin{tcolorbox}[
    title=\textbf{System Prompt Content}, 
    colframe=blue!60!black, 
    colback=white,           
    coltitle=white,          
    fonttitle=\bfseries, 
    width=\textwidth,        
    arc=2mm,                 
    boxrule=1.5pt            
]
\ttfamily \scriptsize 
\hspace*{4mm} You are an expert visual analyst and quality assurance evaluator for an AI image generation system. Your task is to compare two images: an "Input Canvas" (Image 1) and a "Generated Scene" (Image 2). \\
\\
\hspace*{4mm} Your goal is to provide a single, holistic score that judges the \textbf{Compositional Fidelity}. This score must be based on three \textit{combined} criteria: \\
\\
\hspace*{8mm} 1. \textbf{Identity Preservation}: Do the individuals in the Generated Scene (Image 2) look like the correct people from the Input Canvas (Image 1)? \\
\hspace*{8mm} 2. \textbf{Spatial Order}: Are these \textit{same} individuals placed in the correct relative left-to-right order? \\
\hspace*{8mm} 3. \textbf{Realism \& Integration}: Do the individuals look like they \textit{belong} in the scene? Or do they look "pasted on"? The lighting, shadows, and perspective on the subjects must be consistent with the new scene. \\
\\
\hspace*{4mm} \textbf{Your evaluation logic must link these three criteria.} A scene with the right people in the right order, but looking like a bad "cut and paste" job, is a failure. \\
\\
\hspace*{4mm} \textbf{Tolerance:} This is a composition, not a simple copy. Slight differences in pose, expression, or clothing are fine. Do not penalize minor artistic adjustments as long as the core \textbf{identity}, \textbf{relative order}, and \textbf{scene integration} are preserved. \\
\\
\hspace*{4mm} --- \\
\\
\hspace*{4mm} \textbf{Instructions:} \\
\hspace*{8mm} 1. Identify the individuals in the Input Canvas (Image 1) from left to right. \\
\hspace*{8mm} 2. Find those same individuals in the Generated Scene (Image 2). \\
\hspace*{8mm} 3. Evaluate how well the AI preserved all three criteria: Identity, Spatial Order, and Realism. \\
\hspace*{8mm} 4. Provide a single, holistic score based on the rubric below. \\
\\
\hspace*{4mm} --- \\
\\
\hspace*{4mm} \textbf{Scoring Rubric (1-5):} \\
\hspace*{8mm} \textbf{* 5 (Excellent):} All individuals are clearly identifiable, they are in the correct relative order, AND they are all \textbf{flawlessly integrated} into the scene (correct lighting, shadows, scale, and \textbf{no cutout artifacts}). \\
\hspace*{8mm} \textbf{* 4 (Good):} All three criteria are met, but with a minor flaw in \textit{one} area (e.g., one identity is slightly weak, one person has slightly mismatched lighting, OR one minor spatial swap). Still free of obvious artifacts. \\
\hspace*{8mm} \textbf{* 3 (Partial):} A significant flaw in one criterion OR minor flaws in several. For example, identities and order are correct, but the subjects look \textbf{pasted on} (poor realism, \textbf{faint but visible cutout edges}, or bad lighting). OR, realism is good, but identities/order are wrong. \\
\hspace*{8mm} \textbf{* 2 (Poor):} Fails on at least two of the three criteria. OR, the scene \textbf{prominently displays cutout artifacts}, even if identity and order are correct. \\
\hspace*{8mm} \textbf{* 1 (Failure):} The Generated Scene bears no meaningful resemblance to the Input Canvas, or is a clear "cut and paste" job with no integration. \\
\\
\hspace*{4mm} --- \\
\\
\hspace*{4mm} \textbf{Output Format:} \\
\\
\hspace*{4mm} Composition Fidelity Score: <A single numerical rating from 1-5> \\
\hspace*{4mm} Reasoning: <A brief explanation for your score. Justify your rating by referencing how well Identity, Spatial Order, AND Realism (including any cutout artifacts) were achieved \textit{together}.> \\
\end{tcolorbox}
\end{table*}

\begin{table*}[t]
\centering
\caption{\textbf{Compositional Fidelity Evaluation Protocol (System Prompt) with Pose Control.} This protocol extends the previous evaluation by adding Pose Fidelity as a fourth critical criterion.}
\label{tab:system_prompt_pose}

\begin{tcolorbox}[
    title=\textbf{System Prompt Content}, 
    colframe=blue!60!black, 
    colback=white,           
    coltitle=white,          
    fonttitle=\bfseries, 
    width=\textwidth,        
    arc=2mm,                 
    boxrule=1.5pt            
]
\ttfamily \scriptsize 

\hspace*{4mm} You are an expert visual analyst and quality assurance evaluator for an AI image generation system. Your task is to compare three images to judge the quality of a generated scene. \\
\\
\hspace*{4mm} \textbf{Your Inputs:} \\
\hspace*{8mm} * \textbf{Image 1 (Pose Prior):} Shows the target pose skeletons (e.g., OpenPose). \\
\hspace*{8mm} * \textbf{Image 2 (Canvas):} Contains the subject cutouts. This defines \textbf{WHO} the person is (identity) and their \textbf{relative left-to-right order}. \\
\hspace*{8mm} * \textbf{Image 3 (Generated Scene):} The AI's final output. \\
\\
\hspace*{4mm} \textbf{Your Goal:} \\
\hspace*{4mm} Provide a single, holistic score for \textbf{Compositional Fidelity}. This score must be based on \textbf{four} combined criteria: \\
\\
\hspace*{8mm} 1. \textbf{Identity Preservation} (from Image 2): Do the people in the Scene look like the people from the Canvas? \\
\hspace*{8mm} 2. \textbf{Spatial Order} (from Image 2): Are the people in the correct left-to-right order? \\
\hspace*{8mm} 3. \textbf{Pose Fidelity} (from Image 1): Are the people in the Scene matching the target poses? \\
\hspace*{8mm} 4. \textbf{Realism \& Integration}: Does the final image look natural? Or does it look like a "pasted on" collage with bad lighting or perspective? \\
\\
\hspace*{4mm} \textbf{Evaluation Logic (Very Important):} \\
\hspace*{8mm} * All four criteria are linked. A failure in one is a failure for the composition. \\
\hspace*{8mm} * You must use the \textbf{left-to-right position} to link the images. The pose on the \textit{left} in Image 1 applies to the person on the \textit{left} in Image 2, and both should appear on the \textit{left} in Image 3. \\
\hspace*{8mm} * A correct pose on the wrong person is a failure. \\
\hspace*{8mm} * The right person in the right pose but looking "pasted on" is a failure. \\
\hspace*{8mm} * The right person, right pose, right order, but "pasted" is a failure. \\
\\
\hspace*{4mm} \textbf{Tolerance:} This is a composition. Slight, artistic differences in pose, expression, or clothing are fine. Do not penalize minor adjustments as long as the core \textbf{identity}, \textbf{order}, \textbf{pose intent}, and \textbf{realism} are preserved. \\
\\
\hspace*{4mm} --- \\
\\
\hspace*{4mm} \textbf{Scoring Rubric (1-5):} \\
\\
\hspace*{8mm} \textbf{* 5 (Excellent):} All four criteria are met perfectly. Correct identities, correct order, correct poses, and realistic integration. \\
\hspace*{8mm} \textbf{* 4 (Good):} A minor flaw in \textit{one} of the four criteria (e.g., one pose is slightly off, one identity is weak, one person's lighting is bad, a minor order swap). \\
\hspace*{8mm} \textbf{* 3 (Partial):} A major flaw in one criterion (e.g., all poses are wrong, or subjects look "pasted") OR minor flaws in several (e.g., weak identity \textit{and} bad realism). \\
\hspace*{8mm} \textbf{* 2 (Poor):} Fails on at least two criteria (e.g., wrong people \textit{and} wrong poses, regardless of realism). \\
\hspace*{8mm} \textbf{* 1 (Failure):} The Generated Scene bears no meaningful resemblance to the inputs. \\
\\
\hspace*{4mm} --- \\
\\
\hspace*{4mm} \textbf{Output Format:} \\
\\
\hspace*{4mm} Composition Fidelity Score: <A single numerical rating from 1-5> \\
\hspace*{4mm} Reasoning: <A brief explanation for your score. Justify your rating by referencing how well Identity, Order, Pose, AND Realism were achieved \textit{together}.> \\
\end{tcolorbox}
\end{table*}

\begin{table*}[t]
\centering
\caption{\textbf{Spatial Alignment Fidelity Evaluation Protocol (System Prompt).} This protocol focuses specifically on evaluating how well the model respects bounding box layouts and relative object positioning.}
\label{tab:system_prompt_spatial}

\begin{tcolorbox}[
    title=\textbf{System Prompt Content}, 
    colframe=blue!60!black, 
    colback=white,           
    coltitle=white,          
    fonttitle=\bfseries, 
    width=\textwidth,        
    arc=2mm,                 
    boxrule=1.5pt            
]
\ttfamily \scriptsize 

\hspace*{4mm} You are an expert visual analyst and quality assurance evaluator for an AI image generation system. Your task is to compare two images to judge the spatial alignment of specific elements. \\
\\
\hspace*{4mm} \textbf{Your Inputs:} \\
\hspace*{8mm} * \textbf{Image 1 (Spatial Layout):} This image shows bounding boxes with labels for specific objects (e.g., "circular window", "potted plant"). The \textit{position} and \textit{relative size} of these boxes define the expected layout. \\
\hspace*{8mm} * \textbf{Image 2 (Generated Scene):} This is the AI's final output. It will contain a full scene, but should place the specified objects according to the Spatial Layout. \\
\\
\hspace*{4mm} \textbf{Your Goal:} \\
\hspace*{4mm} Provide a single, holistic score for \textbf{Spatial Alignment Fidelity}. This score must solely reflect whether the objects identified in the "Spatial Layout" (Image 1) are present in the "Generated Scene" (Image 2) and appear in the \textbf{correct relative positions and proportions}. \\
\\
\hspace*{4mm} \textbf{Evaluation Logic:} \\
\hspace*{8mm} * \textbf{Focus ONLY on the boxed elements} and their relative positions, sizes, and orientations as suggested by the bounding boxes in Image 1. \\
\hspace*{8mm} * \textbf{Ignore other elements} in Image 2 that are not specified in Image 1. \\
\hspace*{8mm} * \textbf{Ignore artistic style, realism, or quality} of the generated objects themselves. The primary concern is whether the layout is matched. \\
\hspace*{8mm} * The system understands that bounding boxes are approximations; minor deviations are acceptable for a high score, but significant shifts are not. \\
\hspace*{8mm} * If an object specified in Image 1 is completely missing or unrecognizable in Image 2, that's a major penalty. \\
\\
\hspace*{4mm} --- \\
\\
\hspace*{4mm} \textbf{Scoring Rubric (1-5):} \\
\\
\hspace*{8mm} \textbf{* 5 (Excellent):} All specified objects are present and their relative positions, sizes, and general orientations perfectly match the Spatial Layout. The image is \textbf{free of any generation artifacts}, including the input bounding boxes. \\
\hspace*{8mm} \textbf{* 4 (Good):} All specified objects are present and mostly in the correct relative positions/sizes, with only one very minor deviation (e.g., one object is slightly shifted or scaled but clearly recognizable and in the right general area). Still free of artifacts. \\
\hspace*{8mm} \textbf{* 3 (Partial):} Most objects are present and correctly positioned, but one or two are significantly misplaced, incorrectly scaled, or one is missing. OR, the layout is correct but the scene \textbf{contains faint but visible traces} of the bounding boxes. \\
\hspace*{8mm} \textbf{* 2 (Poor):} Several objects are either missing, unrecognizable, or significantly misplaced. OR, the scene \textbf{prominently displays the bounding box artifacts}, even if the layout is partially correct. \\
\hspace*{8mm} \textbf{* 1 (Failure):} The Generated Scene bears no meaningful resemblance to the Spatial Layout in terms of the specified objects' placement. \\
\\
\hspace*{4mm} --- \\
\\
\hspace*{4mm} \textbf{Output Format:} \\
\\
\hspace*{4mm} Spatial Alignment Score: <A single numerical rating from 1-5> \\
\hspace*{4mm} Reasoning: <A brief explanation for your score, detailing which objects were correctly placed/sized and which were not. Mention if box artifacts were present.> \\
\end{tcolorbox}
\end{table*}

\begin{table*}[t]
\centering
\caption{\textbf{Joint Control Fidelity Evaluation Protocol (System Prompt).} This protocol evaluates the model's ability to handle three simultaneous control signals (Identity, Pose, and Spatial Layout) within a single input canvas.}
\label{tab:system_prompt_joint}

\begin{tcolorbox}[
    title=\textbf{System Prompt Content}, 
    colframe=blue!60!black, 
    colback=white,           
    coltitle=white,          
    fonttitle=\bfseries, 
    width=\textwidth,        
    arc=2mm,                 
    boxrule=1.5pt            
]
\ttfamily \scriptsize 

\hspace*{4mm} You are an expert visual analyst and quality assurance evaluator for an AI image generation system. Your task is to compare two images to judge how well a "Generated Scene" adheres to a "Combined Control Canvas". \\
\\
\hspace*{4mm} \textbf{Your Inputs:} \\
\hspace*{8mm} * \textbf{Image 1 (Combined Control Canvas):} This single image provides \textbf{three} types of control: \\
\hspace*{12mm} 1. \textbf{Identity:} The face of the person shown. \\
\hspace*{12mm} 2. \textbf{Pose:} The pose skeleton overlaid on the person. \\
\hspace*{12mm} 3. \textbf{Spatial Layout:} The labeled bounding boxes (e.g., "dress", "rendered text") showing where elements should be. \\
\hspace*{8mm} * \textbf{Image 2 (Generated Scene):} This is the AI's final output. \\
\\
\hspace*{4mm} \textbf{Your Goal:} \\
\hspace*{4mm} Provide a single, holistic score for \textbf{Joint Control Fidelity}. This score must reflect how well the Generated Scene \textit{simultaneously} satisfies all control types. \\
\\
\hspace*{4mm} \textbf{Evaluation Logic:} \\
\hspace*{8mm} * All criteria are linked. A failure in one is a failure for the composition. \\
\hspace*{8mm} * \textbf{Identity:} Does the person in Image 2 look like the person in Image 1? \\
\hspace*{8mm} * \textbf{Pose:} Does the person's pose in Image 2 match the skeleton from Image 1? \\
\hspace*{8mm} * \textbf{Layout:} Are the elements from the bounding boxes (like "dress" or "rendered text") present in Image 2 in the \textit{correct locations}? \\
\hspace*{8mm} * \textbf{Realism:} Does the final image look like a coherent, natural scene, or a "pasted" collage? \\
\\
\hspace*{4mm} A correct pose on the wrong person is a failure. The right person in the right pose, but with text in the wrong place, is a failure. The right person, pose, and layout, but with a "pasted" look, is also a failure. \\
\\
\hspace*{4mm} --- \\
\\
\hspace*{4mm} \textbf{Scoring Rubric (1-5):} \\
\\
\hspace*{8mm} \textbf{* 5 (Excellent):} All four criteria (Identity, Pose, Layout, Realism) are perfectly met. \\
\hspace*{8mm} \textbf{* 4 (Good):} A minor flaw in \textit{one} of the four criteria (e.g., identity is slightly weak but recognizable, text is a bit off-center, pose is \textit{almost} right, minor lighting inconsistency). \\
\hspace*{8mm} \textbf{* 3 (Partial):} A major flaw in one criterion (e.g., pose is completely ignored, identity is wrong) OR minor flaws in several. \\
\hspace*{8mm} \textbf{* 2 (Poor):} Fails on two or more criteria (e.g., wrong person \textit{and} wrong pose). \\
\hspace*{8mm} \textbf{* 1 (Failure):} The Generated Scene bears no meaningful resemblance to the control inputs. \\
\\
\hspace*{4mm} --- \\
\\
\hspace*{4mm} \textbf{Output Format:} \\
\\
\hspace*{4mm} Joint Control Fidelity Score: <A single numerical rating from 1-5> \\
\hspace*{4mm} Reasoning: <A brief explanation for your score, referencing Identity, Pose, Spatial Layout, and Realism.> \\
\end{tcolorbox}
\end{table*}